\pdfoutput=1
\documentclass[11pt]{article}

\usepackage[preprint]{acl}

\usepackage{times}
\usepackage{latexsym}
\usepackage{tabularx}
\usepackage{enumitem}

\usepackage{booktabs, multirow} % for borders and merged ranges
\usepackage{soul}% for underlines
\usepackage{changepage,threeparttable} % for wide tables
\usepackage{amsfonts,amsmath,amssymb,amsthm}
\usepackage{todonotes}
\usepackage{array}
\newcolumntype{Z}[1]{>{\centering\arraybackslash}p{#1}}

\usepackage[export]{adjustbox}

\usepackage{colortbl}

\usepackage{makecell}

\usepackage{soul}
\usepackage{color}

\usepackage[normalem]{ulem} % [normalem] prevents the package from changing the default behavior of `\emph` to underline.

\usepackage{bm}

\usepackage{pifont}% http://ctan.org/pkg/pifont

\usepackage{url}

\definecolor{col_a}{HTML}{EAC2CA}
\definecolor{col_b}{HTML}{C0D8D3}
\definecolor{col_c}{HTML}{C7AFD4}
\definecolor{col_d}{HTML}{B8D9F4}
\definecolor{col_e}{HTML}{DBD9DB}
\definecolor{col_f}{HTML}{F0C594}
\definecolor{col_g}{HTML}{FEEE86}
\definecolor{col_h}{HTML}{8EC76B}

\usepackage[T1]{fontenc}
\usepackage[utf8]{inputenc}

\usepackage{microtype}

\usepackage{inconsolata}

\usepackage{graphicx}

\title{The Impact of Auxiliary Patient Data on Automated Chest X-Ray Report Generation and How to Incorporate It}

\author{Aaron Nicolson, Shengyao Zhuang, Jason Dowling, \& Bevan Koopman \\
  Australian e-Health Research Centre, CSIRO Health and Biosecurity, Brisbane, Australia \\
  \texttt{aaron.nicolson@csiro.au} \\
}

\begin{document}
\maketitle
\begin{abstract}
This study investigates the integration of diverse patient data sources into multimodal language models for automated chest X-ray (CXR) report generation. Traditionally, CXR report generation relies solely on CXR images and limited radiology data, overlooking valuable information from patient health records, particularly from emergency departments. Utilising the MIMIC-CXR and MIMIC-IV-ED datasets, we incorporate detailed patient information such as vital signs, medicines, and clinical history to enhance diagnostic accuracy. We introduce a novel approach to transform these heterogeneous data sources into embeddings that prompt a multimodal language model; this significantly enhances the diagnostic accuracy of generated radiology reports. Our comprehensive evaluation demonstrates the benefits of using a broader set of patient data, underscoring the potential for enhanced diagnostic capabilities and better patient outcomes through the integration of multimodal data in CXR report generation.
\end{abstract}

\section{Introduction}

\begin{figure}
    \centering
    \includegraphics[scale=0.585, trim=0cm 0cm 0cm 0cm, clip]{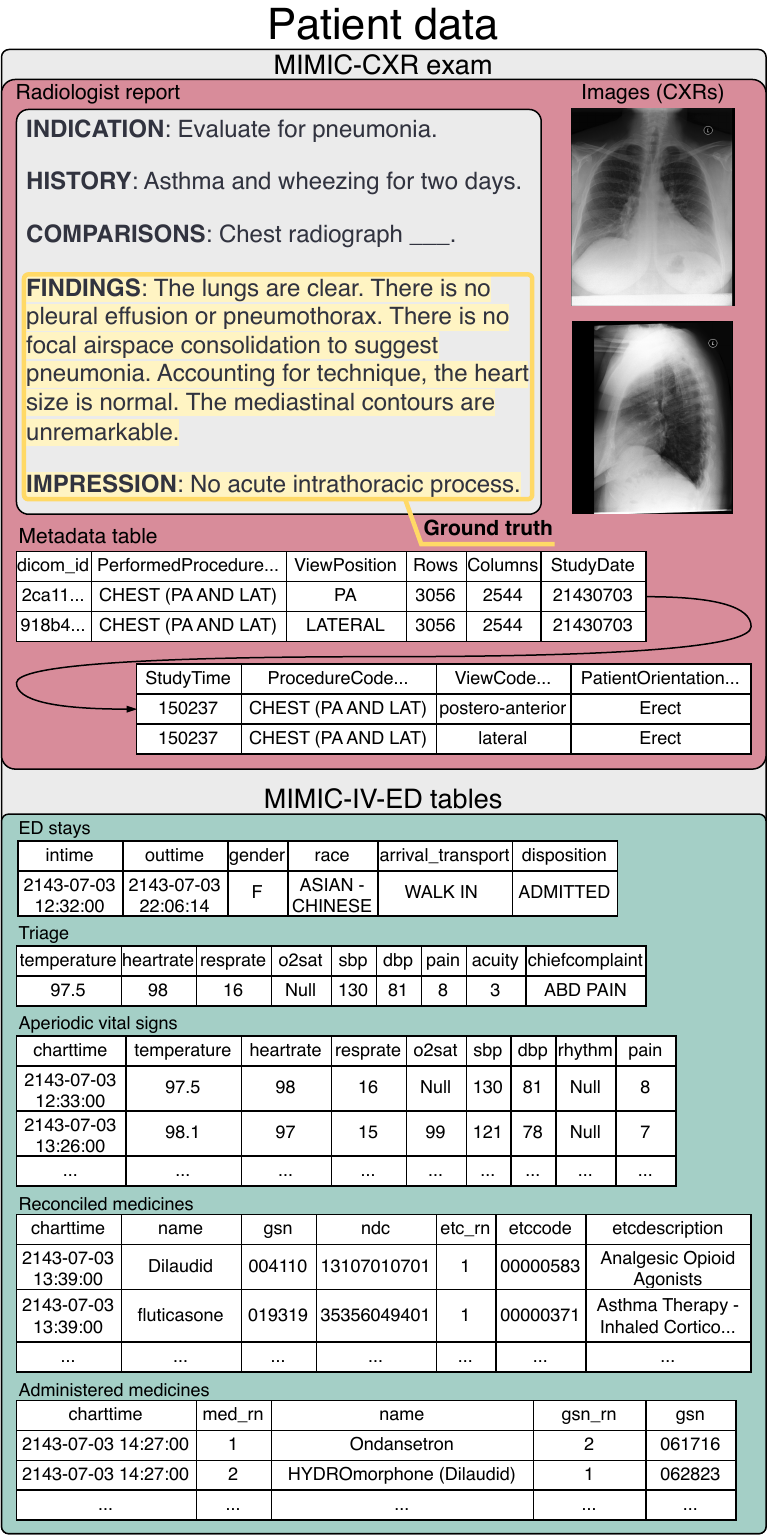}
    \caption{\label{fig:data}The patient data from MIMIC-IV-ED associated with a CXR exam from MIMIC-CXR. This includes the exam's images, the corresponding radiology report, and the associated image metadata. The findings and impression sections of the radiology report form the ground truth for CXR report generation. Emergency-specific data, such as reconciled medicines and aperiodic vital signs, are also available for the patient.}
\end{figure}

Chest X-ray (CXR) exams, which consist of multiple images captured during an imaging session, are essential for diagnosing and managing a wide range of conditions, playing a significant role in patient care. Radiologists interpret these exams and produce a written report with their findings. However, timely reporting is hindered by a multitude of issues, including high patient volumes and limited availability of radiologists~\citep{bailey_understanding_2022}.

Automated CXR report generation using multimodal language models is a promising solution \citep{jones_chest_2021}. Potential benefits include enhanced radiologist effectiveness, streamlining report writing, and improved patient outcomes \citep{shen_grand_2021, irmici_chest_2023}. Early methods produced a separate report for each image within an exam~\cite{Wang_2018}. Later methods improved on this by considering all images of an exam to generate a single report~\cite{miura_improving_2021,nicolson_longitudinal_2024}, and incorporating prior exams for a patient~\cite{wu_deltanet_2022,nicolson_longitudinal_2024}. Including the reason for the exam (the \textit{indication} section in Figure~\ref{fig:data}) offered a further improvement~\cite{nguyen_pragmatic_2023}. This indicates that CXR report generation benefits from the inclusion of a more comprehensive set of patient data.

Incorporating clinical information, including electronic health record (EHR) data, enhanced the interpretation accuracy, clinical relevance, and reporting confidence of radiologists’ findings \cite{castillo_effect_2021}. A growing push to integrate EHR systems into radiology workflows highlights the potential for CXR report generation models to leverage patient data directly \cite{geeslin_electronic_2016}. In this study, we aim to empirically investigate if such data can also improve CXR report generation. To facilitate this, we combine CXR exams from MIMIC-CXR \citep{johnson_mimic-cxr-jpg_2019} with emergency department (ED) patient records from MIMIC-IV-ED \citep{johnson2023mimic}. This provides a wide variety of multimodal data per exam, as shown in Figure \ref{fig:data}. From MIMIC-CXR, we utilise the images, their metadata, and several sections of the radiology report. Notably, we incorporate the history or comparison section of the report, which has not been investigated previously. From MIMIC-IV-ED, we incorporate triage data, aperiodic vital signs, medicines, and other data to provide a wider clinical context. 

We also investigate how to harmonise these heterogeneous data into patient data embeddings to prompt a multimodal language model. In doing so, we develop methods to transform tabular and aperiodic time series data into embeddings that can be used alongside token and image embeddings. We evaluate our model using metrics shown to closely correlate with radiologists’ assessments of reporting \citep{Yu2022.08.30.22279318}. Through our evaluation, we demonstrate that complementary information from different data sources can improve the diagnostic accuracy of CXR report generation. The main contributions of this work are:
\vspace{-5pt}
\begin{itemize}[noitemsep, left=0pt]
    \item An investigation demonstrating how integrating specific patient data sources, such as medicines and vital signs, enhances CXR report generation and improves diagnostic accuracy.
    \item Introducing methods to convert numerical, categorical, text, temporal, and image data into embeddings for a multimodal language model. 
    \item A dataset linking MIMIC-CXR exams with MIMIC-IV-ED records, along with the code and Hugging Face model (available at: \url{https://anonymous.4open.science/r/anon-D83E}).
\end{itemize}

\section{Background and Related Work}

% Some background:
Incorporating more patient data has improved diagnostic accuracy in radiology reporting. Initial improvements came from using multiple images per exam, like EMNLI; CXR exams often include complementary frontal and lateral views of the patient \cite{miura_improving_2021, gaber_lateral_2005}. Methods such as CXRMate enhance diagnostic accuracy by incorporating a patient’s prior exams to identify changes over time \cite{nicolson_longitudinal_2024, wu_deltanet_2022, kellyChestRadiograph2012, bannur_learning_2023, hou-etal-2023-recap}. Including the \textit{indication} section of the radiology report to provide clinical context also provides an improvement \cite{nguyen_pragmatic_2023}. Our investigation focuses on leveraging a more comprehensive set of patient data to improve diagnostic accuracy.

% Initial improvements came from using multiple images per exam, like EMNLI, which often includes complementary frontal and lateral views \cite{miura_improving_2021, gaber_lateral_2005}. Methods such as CXRMate enhance diagnostic accuracy by incorporating a patient’s prior exams to identify changes over time \cite{nicolson_longitudinal_2024, wu_deltanet_2022, kellyChestRadiograph2012}. Including the \textit{indication} section of the radiology report, providing clinical context, also boosts accuracy \cite{nguyen_pragmatic_2023}. This work explores incorporating ED records with CXR exam data for further improvements.

% This trend indicates that improvements can be achieved by providing the model with more comprehensive patient data. In this work, we explore the promising yet unexplored area of incorporating ED records along with CXR exam data.

%This trend indicates that improvements can be achieved by providing the model with more comprehensive patient data. In this work, we delve into the promising yet unexplored area of incorporating patient data from ED records, along with data from CXR exams.

% ED data.
ED records contain a wide range of data, as shown in Figure \ref{fig:data}. The reconciled medicines may include furosemide, a diuretic commonly prescribed for managing fluid overload, often associated with conditions such as pulmonary edema or congestive heart failure. Elevated blood pressure and an increased heart rate in a patient’s vital signs may correlate with findings such as cardiomegaly or vascular changes. Vital signs such as high temperature, elevated respiratory rate, and low oxygen saturation, along with chief complaints of cough and shortness of breath, may suggest pneumonia. Incorporating such data could complement imaging evidence and provide additional context to support better predictions. Our findings demonstrate that ED patient data can indeed improve CXR report generation.

% MIMIC-IV/III.
Recent advancements in integrating multimodal patient data have improved diagnostic and predictive healthcare tasks. A Transformer encoder combining imaging and non-imaging data outperformed single-modality models in diagnosing multiple conditions \cite{khader_multimodal_2023}. Similarly, the MeTra architecture, integrating CXRs and clinical parameters, excelled in predicting ICU survival \cite{khader_medical_2023}, and ETHOS, with zero-shot learning, surpassed single-modality models in predicting mortality, ICU length of stay, and readmission rate \cite{renc_transformer-based_2024}. These studies underscore the value of multimodal data, and our work demonstrates its benefits for CXR report generation.

Multi-task learning has enhanced biomedical models by leveraging shared knowledge across tasks. Med-PaLM M, a generalist biomedical model, excels in classification, question answering, VQA, report summarisation, report generation, and genomic variant calling, using diverse modalities like images, text, and genomics, often outperforming specialised models \citep{tu_towards_2024}. Similarly, MIMIC-CXR has been utilised in multi-task learning with models like MedXChat, which integrates instruction-tuning and Stable Diffusion for tasks like CXR report generation, VQA, and report-to-CXR generation, surpassing other LLM multi-task learners \citep{yang_medxchat_2023}. RaDialog combines visual features and pathology findings to generate accurate radiology reports and enable interactive tasks, improving clinical efficacy. CXR-LLaVA, a multimodal LLM, outperformed models such as GPT-4 Vision and Gemini Pro Vision in CXR report generation \cite{lee_cxr-llava_2024}.

Determining the state-of-the-art in CXR report generation is challenging due to model unavailability and limited comparisons with recent methods. The 2024 Shared Task on Large-Scale Radiology Report Generation (RRG24) aimed to address this by benchmarking models on a common leaderboard. The winning model, CXRMate-RRG24 \cite{nicolson_rrg24_2024}, a derivative of CXRMate, emerged as a strong contender for state-of-the-art. In this work, we compare our model to established models (e.g., EMNLI) and recent benchmarks (e.g., CXRMate-RRG24, CXRMate, CXR-LLaVA, MedXChat, and RaDialog). We ensure a fair comparison by using available code or obtaining generated reports directly from the authors. Our evaluation indicates that our model represents a statistically significant improvement over these.

\begin{figure*}
    \centering
    \vspace{-15pt}
    \includegraphics[scale=0.9]{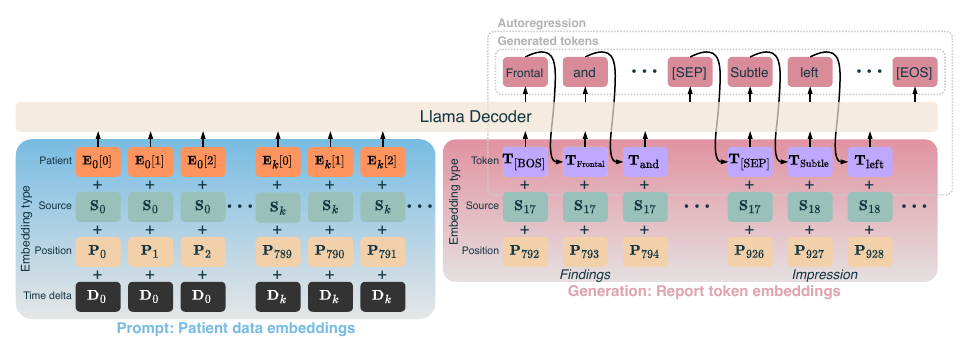}
     \vspace{-5pt}
    \caption{\label{fig:model}Multimodal language model for CXR report generation. The patient data embeddings prompt the decoder to generate the findings and impression sections of a radiology report.}
    \vspace{-10pt}
\end{figure*}

\section{Dataset}\label{sec:dataset}

We construct a dataset of $46\,106$ patients by linking individual patient information from two separate sources: (1) CXR exams from MIMIC-CXR and (2) emergency records from MIMIC-IV-ED. Thus, we consider MIMIC-CXR exams that occurred during an ED stay from MIMIC-IV-ED. Both datasets are publicly available and originate from the Beth Israel Deaconess Medical Center in Boston, MA. 

MIMIC-CXR was formed by first extracting patient identifiers for exams performed in the ED between 2011--2016, and then extracting all exams for this set of patients from all departments between 2011--2016. Each exam includes a semi-structured free-text radiology report (Figure \ref{fig:data}) written by a practising radiologist contemporaneously during routine clinical care. Models are often trained to generate the \textit{findings} and \textit{impression} sections of a radiology report, where the former details the interpretation of a patient's exam and the latter summarises the most important findings. All images and reports were de-identified to protect privacy. Sections from the radiologist reports were extracted using a modification of the official text extraction tool in order to obtain the findings, impression, indication, history, and comparison sections.\footnote{https://anonymous.4open.science/r/anon-D83E} 

MIMIC-IV-ED consists of de-identified data from ED stays between 2011--2019. The data was converted into a denormalised relational database with six primary tables: ED stays, diagnosis, reconciled medicines, administered medicines, triage, and aperiodic vital signs. We do not consider the diagnosis table in this work, as it indicates the outcome of a patient's ED stay. The patients of MIMIC-CXR can be linked to MIMIC-IV-ED via an identifier, allowing an ED-specific dataset to be formed. 

Example tables for a patient's exam are shown in Figure \ref{fig:data}. The dataset was formed by extracting patient exams with times (formed by the `StudyDate' and `StudyTime' columns of the metadata table) that occurred within the `intime' and `outtime' of one of the patient's ED stays.\footnote{Exam \texttt{59128861} was removed as it overlapped with two separate ED stays of a patient.} Events during an ED stay that occurred after the exam were removed. Exams with either a missing findings or impression section were not considered. Using the official splits of MIMIC-CXR, this gave a train/validation/test split of $45\,527$/343/236 patients, $76\,398$/556/958 exams, and $151\,818$/$1\,137$/$1\,812$ CXRs. Further details are provided in Appendix \ref{sec:dataset_details}.

\begin{figure*}[t]
    \centering
    \includegraphics[scale=0.94, trim=1.6cm 0cm 0cm 0cm, clip]{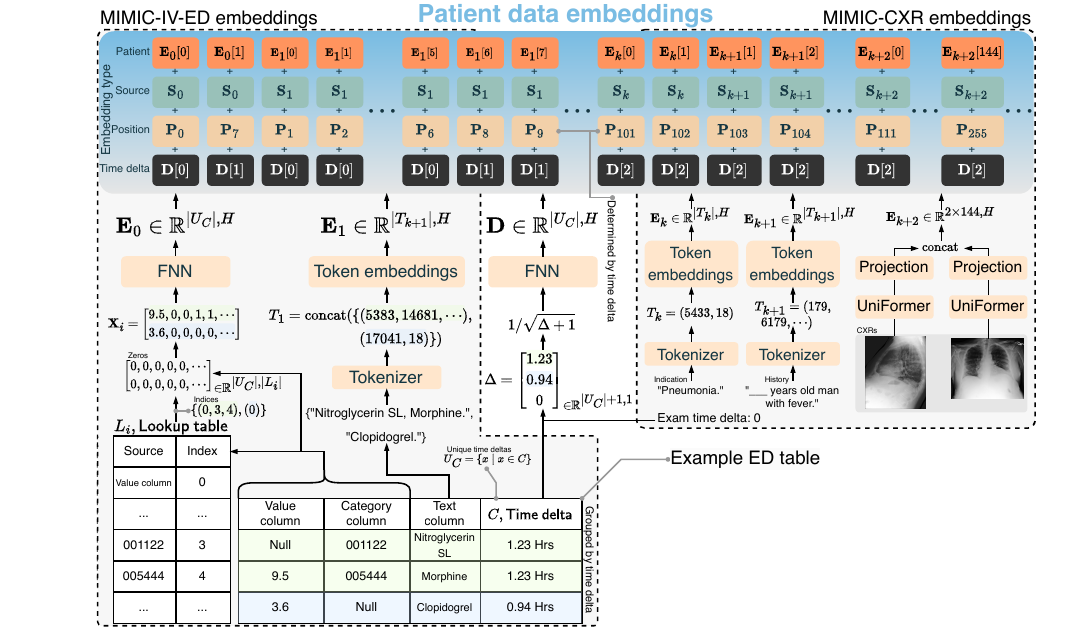}
     \vspace{-5pt}
    \caption{\label{fig:embeddings}Proposed patient data embeddings from the multiple heterogeneous data types taken from MIMIC-IV-ED and MIMIC-CXR. The embeddings are formed from numerical, categorical, textual, temporal, and image data.}
     \vspace{-10pt}
\end{figure*}

%The percentage of studies with a previous study for the training/validation/test splits after dropping studies was $.\%$/$.\%$/$.\%$.

% \citep{DemnerFushman_2015} released a dataset called \textbf{IU X-ray} that consisted of $3{,}955$ English radiology reports and $7{,}470$ CXRs in both DICOM and PNG formats, where both the reports and CXRs were de-identified automatically.
% Each report is associated with a single ... We refer the reader to the survey conducted by \cite[Subsection 2.1]{Pavlopoulos_2021} for a more detailed analysis of MIMIC-CXR and IU X-ray. 
% IU X-ray for testing:
% Note, we cannot test prompt on this because longitudinal information between a patients studies is unavailable.

% \begin{figure*}
%     \centering
%     % left, bottom, right, top.
%     \includegraphics[scale=0.8, trim=1.25cm 0cm 0cm 0cm]{embeddings.pdf}
%     \caption{\label{fig:embeddings}Patient data embeddings.}
% \end{figure*}

\section{Methods} 

    We develop a novel approach to transform different sources of patient data from MIMIC-CXR and MIMIC-IV-ED into embeddings; these are then used to prompt a multimodal language model to generate the findings and impression sections of the radiology report, as illustrated in Figure \ref{fig:model}. Each embedding of the prompt is the summation of a \textit{patient data embedding}, a \textit{source embedding}, a \textit{position embedding}, and a \textit{time delta embedding}. Source embeddings differentiate the source of the datum, for example, the `chief complaint' column of the triage table, the indication section, or an image. A time delta embedding represents the time difference between an event and the exam. The patient data embeddings originate from three main groups: the tables of MIMIC-IV-ED; the report, images, and metadata of the current exam from MIMIC-CXR; and the patient’s prior exams (also originating from MIMIC-CXR). The prior exam and image embeddings are described in Section \ref{sec:appendix_embeddings} and Subsection \ref{sec:appendix_model}, respectively.

\subsection{Time, Position, \& Source Embeddings}

% The ED information from MIMIC-IV-ED is typically recorded as discrete events, such as medicines administered or vital signs measured, each with a specific timestamp. 

% Events that occur closer to the time of the patient's exam are generally more relevant for diagnostic purposes. To capture this, a time delta is calculated by subtracting the time of an event from the time of the exam. The exam time originates from MIMIC-CXR's metadata table, whereas most of the MIMIC-IV-ED tables have event times for each row. 

% As shown in Figure \ref{fig:embeddings}, the time delta is first converted to hours and then mapped using $1/\sqrt{\bm{\Delta}+1}$, assigning higher weights to events that occurred closer to the exam. The mapped time deltas are then passed through a feedforward neural network (FNN) defined as \( f(\bm{\Delta} \bm{W}_{\text{1}}) \bm{W}_{\text{2}} \), where \( \bm{W}_{\text{1}} \in \mathbb{R}^{1,2048} \), \( \bm{W}_{\text{2}} \in \mathbb{R}^{2048,H} \), \( f(\cdot) \) is the sigmoid linear unit (SiLU) activation function \citep{hendrycks_gaussian_2016}, and \( H \) is the hidden size of the decoder. This process generates the time delta embeddings, which are subsequently added to the embeddings of their respective sources. As shown in Figure \ref{fig:model}, time delta embeddings are only applied to the prompt. Patient data from the current exam, such as the images, have a time delta of zero, while data from prior exams have a positive time delta.

Events from MIMIC-IV-ED, e.g., administered medicines, are timestamped and are more relevant as they occur closer to the exam time \cite{abacha_investigation_nodate}. Hence, time delta embeddings are used to indicate this to the model. The time delta is the event time subtracted from the exam time, converted to hours and mapped using $\bm{D}=1/\sqrt{\bm{\Delta}+1}$, emphasising recent events. These mapped time deltas are processed via a feedforward neural network (FNN), \(  f(\bm{D} \bm{W}_{\text{1}}) \bm{W}_{\text{2}} \), where \( \bm{W}_{\text{1}} \in \mathbb{R}^{1,2048} \), \( \bm{W}_{\text{2}} \in \mathbb{R}^{2048,H} \), \( f(\cdot) \) is the SiLU activation \citep{hendrycks_gaussian_2016}, and \( H \) is the decoder's hidden size. As shown in Figure \ref{fig:model}, these embeddings are applied only to the prompt. The exam time for current and prior exam data was used as the event time for the time delta calculation.

The position embeddings are ordered by the time delta (Figure \ref{fig:embeddings}). This is due to the rotary position embeddings of the decoder; tokens that are closer together are given more importance. Hence, the smaller the time delta, the closer the patient data embedding's position is to the report token embeddings. Following \citet{nicolson_longitudinal_2024}, each unique patient data source is given its own source embedding. This includes the images, each report section, each table's text column and value-category columns (described in the next section), prior images, and prior report sections.

% \begin{figure}
%     \centering
%     \includegraphics[scale=0.925, trim=0.75cm 0cm 0cm 0cm, clip]{cxr_embeddings.drawio.pdf}
%     \caption{\label{fig:cxr_embeddings}Embeddings from MIMIC-CXR sources.}
% \end{figure}

\subsection{Patient Data Embeddings: Tabular Data}

An example table and its conversion to embeddings is shown in Figure \ref{fig:embeddings}.  The columns of each table were designated as value, category, text, or time columns. Value columns contained numeric data, while category columns contained categorical data. To convert an exam's tabular data to embeddings, data from value and category columns were grouped by their time delta, where each group formed a feature vector. The feature vector initially consisted of zeros. Values and categories from the group were then used to set its values based on indices determined by a lookup table. For value columns, the lookup table determined the index where the numeric value was placed. For category columns, it determined which indices were activated (set to 1).

Next, the feature vector was passed through an FNN $f( \bm{X}_i \bm{W}_{\text{1}} ) \bm{W}_{\text{2}}$ to form the embedding, where $\bm{X}_i \in \mathbb{R}^{|U_C|,|L_i|}$ are the grouped features, $U_C$ is the set of unique time deltas, $\bm{W}_{\text{1}} \in \mathbb{R}^{|L_i|,2048}$ and $\bm{W}_{\text{2}} \in \mathbb{R}^{2048,H}$, $L_i$ is a lookup table, and $i$ designates the table. Each table has a unique FNN and lookup table. Rows for a value column always had a unique time, preventing multiple values from the same column in a group. We investigated alternatives to form the value-category embeddings in Section \ref{sec:results}. The described framework was found to be the most efficient. Columns with a high cardinality were set as text columns. Text embeddings were formed via the decoder's tokenizer and token embeddings. Text embeddings were given the time delta embedding from their respective row. The column designation for each table in Figure \ref{fig:data} is described in the Appendix \ref{sec:appendix_columns}.

\subsection{Patient Data Embeddings: Report Sections}

Here, we consider five sections of the radiology report: the findings, impression, indication, history, and comparison sections. The findings and impression sections serve as the ground truth to be generated. The remainder form part of the patient data embeddings. The indication section explains the reason for the exam, such as symptoms or suspected conditions. The history section provides relevant medical history, such as past conditions and treatments. The comparison section mentions any prior exams, which are used to capture disease progression. These sections provide context that guides the interpretation of the exam, influencing the content of the findings and impression sections. The embeddings were formed via the decoder's tokenizer and token embeddings. Of these, the history and comparison sections have not been investigated for CXR report generation. The comparison section was used only when prior exams were considered.

\begin{table*}[t]
\centering
\caption{\label{tab:main_results}Results of the various patient data sources on the test set described in Section \ref{sec:dataset}. Results were calculated over ten training runs ($n=9\,580 ~{\rm exams}; ~958 \times 10~{\rm runs}$). \underline{Underlined} and \(\dashuline{\text{dashed underlined}}\) scores indicate a significant difference to the scores of `Images' and `Images + effective sources ($h=0$)', respectively ($p<0.05$). Evaluation is performed on both the \textbf{findings} and \textbf{impression} sections. 
}
\vspace{-10pt}

% \vspace{-5pt}
\small
\begin{tabular}{@{}lllllllllll@{}}
\toprule

Patient data sources & RG & CX & CB & G & BS & R-L & B4 & $\overline{|\bm{\mathcal{E}}[:,0]|}$ \\

\midrule
\multicolumn{9}{c}{\cellcolor[RGB]{200,200,200} \textit{Images only}} \\
Images & \cellcolor[RGB]{243,236,217} 24.54 & \cellcolor[RGB]{234,237,242} 30.10 & \cellcolor[RGB]{243,239,234} 59.25 & \cellcolor[RGB]{212,226,239} 35.16 & \cellcolor[RGB]{245,229,182} 24.26 & \cellcolor[RGB]{211,225,239} 25.91 & \cellcolor[RGB]{243,238,229} 4.75 & \cellcolor[RGB]{239,240,242} 272.4\\
\multicolumn{9}{c}{\cellcolor[RGB]{200,200,200} \textit{Patient emergency department data (MIMIC-IV-ED)}} \\
Images + ED stays & \cellcolor[RGB]{242,240,236} 24.20 & \cellcolor[RGB]{239,240,242} 29.55 & \cellcolor[RGB]{242,240,237} 58.37 & \cellcolor[RGB]{236,238,242} 34.64 & \cellcolor[RGB]{244,233,203} 24.06 & \cellcolor[RGB]{224,232,241} 25.77 & \cellcolor[RGB]{242,240,236} 4.66 & \cellcolor[RGB]{239,240,242} 273.4\\
Images + triage & \cellcolor[RGB]{244,235,213} 24.59 & \cellcolor[RGB]{180,210,236} 31.33 & \cellcolor[RGB]{247,224,159} \underline{62.79} & \cellcolor[RGB]{152,195,233} 35.78 & \cellcolor[RGB]{246,225,165} 24.40 & \cellcolor[RGB]{205,223,239} 25.96 & \cellcolor[RGB]{243,238,229} 4.76 & \cellcolor[RGB]{239,240,242} 278.9\\
Images + vital signs & \cellcolor[RGB]{242,239,235} 24.23 & \cellcolor[RGB]{219,230,240} 30.61 & \cellcolor[RGB]{243,236,217} 60.61 & \cellcolor[RGB]{212,226,240} 35.15 & \cellcolor[RGB]{244,233,205} 24.04 & \cellcolor[RGB]{216,228,240} 25.86 & \cellcolor[RGB]{243,239,233} 4.70 & \cellcolor[RGB]{239,240,242} 274.7\\
Images + reconciled medicines & \cellcolor[RGB]{248,220,140} 25.10 & \cellcolor[RGB]{121,179,230} 32.05 & \cellcolor[RGB]{251,207,77} \underline{64.70} & \cellcolor[RGB]{72,154,226} \underline{36.32} & \cellcolor[RGB]{249,216,118} 24.71 & \cellcolor[RGB]{154,196,234} 26.29 & \cellcolor[RGB]{244,234,206} 4.93 & \cellcolor[RGB]{237,239,242} 355.6\\
Images + administered medicines & \cellcolor[RGB]{242,239,235} 24.22 & \cellcolor[RGB]{226,233,241} 30.40 & \cellcolor[RGB]{243,237,225} 60.13 & \cellcolor[RGB]{229,235,241} 34.85 & \cellcolor[RGB]{244,234,211} 23.97 & \cellcolor[RGB]{235,238,242} 25.61 & \cellcolor[RGB]{242,240,237} 4.58 & \cellcolor[RGB]{239,240,242} 273.0\\
\multicolumn{9}{c}{\cellcolor[RGB]{200,200,200} \textit{Patient radiology data (MIMIC-CXR)}} \\
Images + indication & \cellcolor[RGB]{247,223,156} 25.01 & \cellcolor[RGB]{43,139,223} 32.78 & \cellcolor[RGB]{253,199,34} \underline{65.49} & \cellcolor[RGB]{138,188,232} 35.88 & \cellcolor[RGB]{249,215,115} 24.73 & \cellcolor[RGB]{148,193,233} 26.32 & \cellcolor[RGB]{247,223,156} 5.15 & \cellcolor[RGB]{239,240,242} 279.5\\
Images + history & \cellcolor[RGB]{246,228,177} 24.88 & \cellcolor[RGB]{155,197,234} 31.66 & \cellcolor[RGB]{249,215,115} \underline{63.91} & \cellcolor[RGB]{154,196,234} 35.76 & \cellcolor[RGB]{251,208,82} \underline{24.91} & \cellcolor[RGB]{58,146,224} \underline{26.70} & \cellcolor[RGB]{255,194,10} \textbf{\underline{5.54}} & \cellcolor[RGB]{239,240,242} 277.0\\
Images + metadata & \cellcolor[RGB]{242,240,237} 24.07 & \cellcolor[RGB]{225,233,241} 30.42 & \cellcolor[RGB]{243,238,229} 59.75 & \cellcolor[RGB]{231,236,241} 34.79 & \cellcolor[RGB]{243,236,219} 23.86 & \cellcolor[RGB]{236,238,242} 25.59 & \cellcolor[RGB]{242,240,237} 4.58 & \cellcolor[RGB]{239,240,242} 273.4\\
\multicolumn{9}{c}{\cellcolor[RGB]{200,200,200} \textit{Prior exams}} \\
Images + $h=1$ & \cellcolor[RGB]{244,232,199} 24.71 & \cellcolor[RGB]{201,220,238} 30.98 & \cellcolor[RGB]{246,225,165} \underline{62.60} & \cellcolor[RGB]{147,193,233} 35.81 & \cellcolor[RGB]{246,226,168} 24.38 & \cellcolor[RGB]{200,220,238} 26.00 & \cellcolor[RGB]{243,237,222} 4.82 & \cellcolor[RGB]{206,223,239} 603.0\\
Images + $h=2$ & \cellcolor[RGB]{244,235,215} 24.56 & \cellcolor[RGB]{173,206,236} 31.43 & \cellcolor[RGB]{245,229,182} \underline{62.09} & \cellcolor[RGB]{189,214,237} 35.43 & \cellcolor[RGB]{244,233,205} 24.04 & \cellcolor[RGB]{222,231,241} 25.80 & \cellcolor[RGB]{243,236,221} 4.84 & \cellcolor[RGB]{128,182,231} 878.1\\
Images + $h=3$ & \cellcolor[RGB]{243,236,221} 24.50 & \cellcolor[RGB]{213,227,240} 30.73 & \cellcolor[RGB]{243,238,228} 59.89 & \cellcolor[RGB]{208,224,239} 35.21 & \cellcolor[RGB]{244,233,205} 24.03 & \cellcolor[RGB]{221,230,240} 25.82 & \cellcolor[RGB]{243,239,233} 4.70 & \cellcolor[RGB]{13,123,220} 1134.3\\
Images + $h=1$ + comparison & \cellcolor[RGB]{246,226,171} 24.92 & \cellcolor[RGB]{171,205,235} 31.46 & \cellcolor[RGB]{247,223,154} \underline{62.93} & \cellcolor[RGB]{144,191,233} 35.84 & \cellcolor[RGB]{246,227,172} 24.34 & \cellcolor[RGB]{196,218,238} 26.03 & \cellcolor[RGB]{244,235,213} 4.89 & \cellcolor[RGB]{205,223,239} 607.4\\
Images + $h=2$ + comparison & \cellcolor[RGB]{243,236,219} 24.52 & \cellcolor[RGB]{199,219,238} 31.01 & \cellcolor[RGB]{244,232,201} \underline{61.36} & \cellcolor[RGB]{227,234,241} 34.89 & \cellcolor[RGB]{243,236,216} 23.90 & \cellcolor[RGB]{234,237,242} 25.62 & \cellcolor[RGB]{243,239,232} 4.72 & \cellcolor[RGB]{127,182,231} 882.6\\
Images + $h=3$ + comparison & \cellcolor[RGB]{243,239,232} 24.31 & \cellcolor[RGB]{204,222,239} 30.93 & \cellcolor[RGB]{243,237,225} 60.10 & \cellcolor[RGB]{239,240,242} 34.35 & \cellcolor[RGB]{242,240,237} \underline{23.31} & \cellcolor[RGB]{239,240,242} \underline{25.39} & \cellcolor[RGB]{243,239,232} 4.72 & \cellcolor[RGB]{12,123,220} \textbf{1138.8}\\
\multicolumn{9}{c}{\cellcolor[RGB]{200,200,200} \textit{All effective sources (triage, reconciled medicines, indication, and history)}} \\
Images + effective sources ($h=0$) & \cellcolor[RGB]{253,201,45} \underline{25.52} & \cellcolor[RGB]{77,156,226} 32.49 & \cellcolor[RGB]{255,194,10} \textbf{\underline{65.93}} & \cellcolor[RGB]{82,159,227} \underline{36.26} & \cellcolor[RGB]{253,198,30} \underline{25.16} & \cellcolor[RGB]{27,130,221} \underline{26.81} & \cellcolor[RGB]{250,211,94} \underline{5.34} & \cellcolor[RGB]{237,239,242} 373.9\\
Images + effective sources ($h=1$) & \cellcolor[RGB]{248,220,139} 25.11 & \cellcolor[RGB]{192,216,238} 31.14 & \cellcolor[RGB]{244,233,205} \underline{61.19} & \cellcolor[RGB]{149,194,233} 35.80 & \cellcolor[RGB]{251,207,75} \underline{24.95} & \cellcolor[RGB]{120,179,230} \underline{26.45} & \cellcolor[RGB]{248,220,139} \underline{5.21} & \cellcolor[RGB]{183,211,237} 704.5\\
Images + effective sources ($h=1$ + comparison) & \cellcolor[RGB]{247,222,149} 25.05 & \cellcolor[RGB]{216,228,240} 30.68 & \cellcolor[RGB]{244,234,210} 60.99 & \cellcolor[RGB]{130,184,231} 35.94 & \cellcolor[RGB]{251,207,76} \underline{24.94} & \cellcolor[RGB]{114,176,230} \underline{26.48} & \cellcolor[RGB]{248,218,129} \underline{5.24} & \cellcolor[RGB]{181,210,236} 709.0\\
\multicolumn{9}{c}{\cellcolor[RGB]{200,200,200} \textit{Ablation from Images + effective sources ($h=0$)}} \\
\quad\quad - triage & \cellcolor[RGB]{255,194,10} \textbf{25.65} & \cellcolor[RGB]{34,134,222} 32.85 & \cellcolor[RGB]{253,200,41} 65.38 & \cellcolor[RGB]{70,153,225} 36.33 & \cellcolor[RGB]{255,194,10} \textbf{25.25} & \cellcolor[RGB]{44,139,223} 26.75 & \cellcolor[RGB]{250,211,98} 5.33 & \cellcolor[RGB]{237,239,242} 367.4\\
\quad\quad - reconciled medicines & \cellcolor[RGB]{251,205,68} 25.43 & \cellcolor[RGB]{78,157,226} 32.48 & \cellcolor[RGB]{254,197,26} 65.63 & \cellcolor[RGB]{54,145,224} 36.42 & \cellcolor[RGB]{254,194,14} 25.23 & \cellcolor[RGB]{12,123,220} \textbf{26.86} & \cellcolor[RGB]{251,206,71} 5.40 & \cellcolor[RGB]{239,240,242} 290.7\\
\quad\quad - indication & \cellcolor[RGB]{252,204,60} 25.46 & \cellcolor[RGB]{25,129,221} 32.92 & \cellcolor[RGB]{254,196,23} 65.69 & \cellcolor[RGB]{56,146,224} 36.41 & \cellcolor[RGB]{254,195,18} 25.21 & \cellcolor[RGB]{32,133,222} 26.79 & \cellcolor[RGB]{250,209,86} 5.36 & \cellcolor[RGB]{237,239,242} 366.7\\
\quad\quad - history & \cellcolor[RGB]{251,206,73} 25.41 & \cellcolor[RGB]{72,154,226} 32.53 & \cellcolor[RGB]{254,195,16} 65.82 & \cellcolor[RGB]{12,123,220} \textbf{36.65} & \cellcolor[RGB]{253,199,39} 25.12 & \cellcolor[RGB]{53,144,224} 26.72 & \cellcolor[RGB]{251,208,83} 5.37 & \cellcolor[RGB]{237,239,242} 369.2\\
\quad\quad - time delta & \cellcolor[RGB]{250,211,97} 25.31 & \cellcolor[RGB]{12,123,220} \textbf{33.03} & \cellcolor[RGB]{254,196,22} 65.72 & \cellcolor[RGB]{96,166,228} 36.17 & \cellcolor[RGB]{253,200,43} 25.10 & \cellcolor[RGB]{44,139,223} 26.75 & \cellcolor[RGB]{250,211,94} 5.34 & \cellcolor[RGB]{237,239,242} 373.9\\

\bottomrule
 
\end{tabular}
 \vspace{-15pt}
\end{table*}

\section{Experiment Setup} \label{sec:body_experiment_setup}

Our multimodal language model, illustrated in Figure \ref{fig:model}, is based on CXRMate-RRG24; it features a Llama decoder and the UniFormer as the image encoder. The training procedure for our model involved three stages: (1) initial training on the MIMIC-CXR training set using only images as input with Teacher Forcing (TF) \citep{williams_learning_1989}, (2) further training on the dataset described in Section \ref{fig:data} with the inputs detailed in Table \ref{tab:main_results}, again using TF, and (3) reinforcement learning on the same dataset through self-critical sequence training (SCST) \cite{rennie_self-critical_2017} (only for Table \ref{tab:comparison}). Our evaluation metrics included four that capture the semantics of radiology reporting --- RadGraph-F1 (RG), CheXbert-F1 (CX), CXR-BERT (CB), and GREEN (G) --- as well as three natural language generation metrics: BERTScore-F1 (BS), ROUGE-L (R-L), and BLEU-4 (B4). We also propose a metric that measures \textit{n}-gram repetition rate, namely the absence of repeated \textit{n}-grams (ARN). Comprehensive details on ARN and the other metrics, the model architecture, training procedure, significance testing, and comparison methods are provided in Appendix \ref{sec:appendix_experiment_setup}.

\begin{table*}[t]
\centering
\caption{\label{tab:comparison}Benchmark models on the test set described in Section \ref{sec:dataset} ($n=958$). Evaluation is on the \textbf{findings} section only. \underline{Underlined} indicates statistical significance between the top two scores ($p<0.05$). In the `Train samples' column, `images' means the model generates reports per image, while `exams' means a report generated per exam.}
 \vspace{-10pt}

\small
\setlength{\tabcolsep}{5pt} % Reduce column spacing

\begin{tabular}{@{}lllllllllll@{}}
\toprule
 Model & Train samples &  RG & CX & CB & G & BS & R-L & B4 & ARN \\ \midrule

EMNLI \cite{miura_improving_2021} & 152\,173 exams & \cellcolor[RGB]{250,212,99} 29.1 & \cellcolor[RGB]{205,223,239} 28.9 & \cellcolor[RGB]{247,222,151} 66.6 & \cellcolor[RGB]{16,125,220} 41.5 & \cellcolor[RGB]{242,240,236} 24.4 & \cellcolor[RGB]{171,205,235} 29.3 & \cellcolor[RGB]{243,239,234} 4.1 & \cellcolor[RGB]{238,240,242} 95.1\\
CMN \cite{chen_cross-modal_2021} & 270\,790 images & \cellcolor[RGB]{243,239,232} 23.6 & \cellcolor[RGB]{231,236,241} 24.3 & \cellcolor[RGB]{243,237,224} 49.4 & \cellcolor[RGB]{179,209,236} 36.6 & \cellcolor[RGB]{242,240,237} 19.7 & \cellcolor[RGB]{210,225,239} 27.8 & \cellcolor[RGB]{242,239,235} 4.0 & \cellcolor[RGB]{64,150,225} 99.3\\
TranSQ \cite{wang_transq_2022} & 368\,960 images & \cellcolor[RGB]{249,216,121} 28.7 & \cellcolor[RGB]{190,215,237} 30.4 & \cellcolor[RGB]{245,228,180} 62.3 & \cellcolor[RGB]{143,190,233} 38.2 & \cellcolor[RGB]{242,240,237} 20.4 & \cellcolor[RGB]{239,240,242} 23.3 & \cellcolor[RGB]{243,239,234} 4.1 & \cellcolor[RGB]{143,190,233} 98.5\\
RGRG \cite{tanida_interactive_2023} & 166\,512 images & \cellcolor[RGB]{242,239,235} 22.9 & \cellcolor[RGB]{235,238,242} 22.8 & \cellcolor[RGB]{242,240,236} 37.9 & \cellcolor[RGB]{234,237,242} 31.1 & \cellcolor[RGB]{242,240,237} 23.4 & \cellcolor[RGB]{239,240,242} 22.0 & \cellcolor[RGB]{242,240,236} 3.7 & \cellcolor[RGB]{227,234,241} 96.5\\
CvT2DistilGPT2 \cite{nicolson_improving_2022} & 270\,790 images & \cellcolor[RGB]{243,238,230} 23.9 & \cellcolor[RGB]{201,220,238} 29.3 & \cellcolor[RGB]{245,231,192} 59.8 & \cellcolor[RGB]{171,205,235} 37.0 & \cellcolor[RGB]{242,239,235} 24.8 & \cellcolor[RGB]{192,216,238} 28.6 & \cellcolor[RGB]{243,236,220} 5.4 & \cellcolor[RGB]{98,167,228} 99.0\\
RaDialog \cite{pellegrini_radialog_2023} & 276\,778 images & \cellcolor[RGB]{243,238,228} 24.4 & \cellcolor[RGB]{12,123,220} \textbf{38.4} & \cellcolor[RGB]{245,230,188} 60.7 & \cellcolor[RGB]{206,223,239} 34.9 & \cellcolor[RGB]{243,239,232} 26.2 & \cellcolor[RGB]{225,233,241} 26.7 & \cellcolor[RGB]{243,238,229} 4.8 & \cellcolor[RGB]{239,240,242} 94.4\\
MedXChat \cite{yang_medxchat_2023} & 270\,790 images & \cellcolor[RGB]{242,240,237} 21.0 & \cellcolor[RGB]{239,240,242} 13.1 & \cellcolor[RGB]{242,240,237} 21.3 & \cellcolor[RGB]{232,236,242} 31.4 & \cellcolor[RGB]{242,240,237} 19.3 & \cellcolor[RGB]{239,240,242} 23.8 & \cellcolor[RGB]{242,239,235} 4.0 & \cellcolor[RGB]{182,211,237} 97.9\\
CXR-LLaVA-v2 \cite{lee_cxr-llava_2024} & 193\,513 images & \cellcolor[RGB]{242,240,237} 19.4 & \cellcolor[RGB]{237,239,242} 20.7 & \cellcolor[RGB]{243,239,231} 44.1 & \cellcolor[RGB]{239,240,242} 24.0 & \cellcolor[RGB]{242,240,237} 23.6 & \cellcolor[RGB]{239,240,242} 21.1 & \cellcolor[RGB]{242,240,237} 1.7 & \cellcolor[RGB]{12,123,220} \textbf{99.7}\\
CXRMate \cite{nicolson_longitudinal_2024} & 125\,395 exams & \cellcolor[RGB]{244,232,197} 26.5 & \cellcolor[RGB]{136,187,232} 33.9 & \cellcolor[RGB]{249,214,109} 71.3 & \cellcolor[RGB]{71,153,225} 40.3 & \cellcolor[RGB]{244,233,205} 30.5 & \cellcolor[RGB]{178,208,236} 29.1 & \cellcolor[RGB]{248,218,130} 7.5 & \cellcolor[RGB]{164,201,235} 98.2\\
CXRMate-RRG24 \cite{nicolson_rrg24_2024} & 550\,395 exams & \cellcolor[RGB]{249,214,110} 28.9 & \cellcolor[RGB]{179,209,236} 31.2 & \cellcolor[RGB]{244,232,199} 58.2 & \cellcolor[RGB]{76,156,226} 40.2 & \cellcolor[RGB]{244,232,198} 31.0 & \cellcolor[RGB]{189,214,237} 28.7 & \cellcolor[RGB]{245,229,182} 6.6 & \cellcolor[RGB]{192,216,238} 97.7\\
\midrule
Images + effective sources ($h=0$) & 76\,398 exams & \cellcolor[RGB]{243,236,221} 25.1 & \cellcolor[RGB]{198,219,238} 29.6 & \cellcolor[RGB]{247,223,156} 66.0 & \cellcolor[RGB]{174,206,236} 36.9 & \cellcolor[RGB]{244,235,215} 29.4 & \cellcolor[RGB]{210,225,239} 27.8 & \cellcolor[RGB]{244,234,210} 5.8 & \cellcolor[RGB]{143,190,233} 98.5\\
\quad+ RL (CXR-BERT + BERTScore reward) & 76\,398 exams & \cellcolor[RGB]{255,194,10} \textbf{30.4} & \cellcolor[RGB]{94,165,228} 35.7 & \cellcolor[RGB]{255,194,10} \textbf{\underline{79.1}} & \cellcolor[RGB]{12,123,220} \textbf{41.6} & \cellcolor[RGB]{254,196,24} \underline{37.2} & \cellcolor[RGB]{57,146,224} \underline{31.6} & \cellcolor[RGB]{255,194,10} \textbf{\underline{8.7}} & \cellcolor[RGB]{239,240,242} 93.5\\
\quad\quad + reward per section & 76\,398 exams & \cellcolor[RGB]{253,198,33} 30.1 & \cellcolor[RGB]{139,188,232} 33.7 & \cellcolor[RGB]{254,196,21} \underline{78.3} & \cellcolor[RGB]{12,123,220} \textbf{41.6} & \cellcolor[RGB]{255,194,10} \textbf{\underline{37.5}} & \cellcolor[RGB]{12,123,220} \textbf{\underline{32.2}} & \cellcolor[RGB]{253,201,46} 8.4 & \cellcolor[RGB]{239,240,242} 94.6\\
\quad\quad\quad + ARN reward & 76\,398 exams & \cellcolor[RGB]{254,197,26} 30.2 & \cellcolor[RGB]{141,189,232} 33.6 & \cellcolor[RGB]{254,197,26} \underline{78.0} & \cellcolor[RGB]{54,145,224} 40.7 & \cellcolor[RGB]{254,195,19} \underline{37.3} & \cellcolor[RGB]{35,134,222} \underline{31.9} & \cellcolor[RGB]{249,216,122} 7.6 & \cellcolor[RGB]{64,150,225} 99.3\\

\bottomrule
 
\end{tabular}
\end{table*}

\vspace{-10pt}

\begin{figure*}
    \centering
    % left, bottom, right, top.
    \includegraphics[scale=0.8, trim=0cm 0.1cm 0cm 0cm, clip]{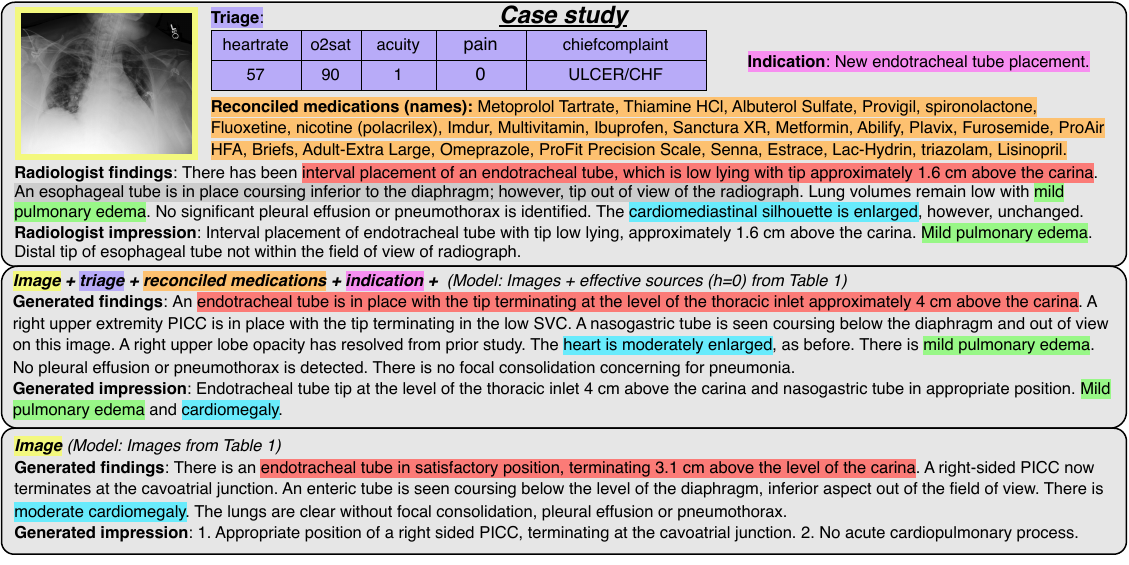}
    \vspace{-10pt}
    \caption{\label{fig:case_study}Case study demonstrating how incorporating auxiliary patient data can aid with report generation.}
    \vspace{-15pt}
\end{figure*}

\section{Results \& Discussion} \label{sec:results}

The impact of different patient data sources on the performance of CXR report generation is summarised in Table \ref{tab:main_results}. This analysis identifies which additional data sources improve performance compared to using only images. Significant improvements were observed by incorporating either triage or reconciled medicine data from MIMIC-IV-ED dataset. Notably, this data markedly improved scores on the radiology report metrics (RG, CX, CB, and G). These findings demonstrate that ED patient data can improve the diagnostic accuracy of CXR report generation. Aperiodic vital sign and administered medicine data did not significantly improve the scores overall, likely due to their frequency of occurrence in the exams (62\% and 37\%, respectively). However, as shown in Table \ref{tab:vitalsign}, a significant improvement in  performance was attained when evaluated solely on exams that include an aperiodic vital sign table.

Incorporating the indication or history section led to significant score improvements. This demonstrates the substantial influence these sections have on the findings and impression sections. Conversely, adding the metadata table did not result in significant score improvements, indicating it lacks valuable information for CXR report generation. While previous studies have established that the indication section boosts CXR report generation \cite{nguyen_pragmatic_2023}, our findings demonstrate that the history section is equally important.

When examining the impact of prior exams, we considered a maximum history size $h$ of up to three, incorporating the findings and impression sections, and images from prior exams. A history size of one or two significantly increased the scores, which is consistent with previous findings \citep{wu_deltanet_2022}. However, performance gradually degraded as the history size increased, which contradicts earlier studies. We suspect this is due to the size of the prompt increasing as \( h \) grows, combined with the limitations of our model architecture. $\overline{|\bm{\mathcal{E}}[:,0]|}$ in Table \ref{tab:main_results} is the average prompt length over the test set, where $\bm{\mathcal{E}} = [\textbf{E}_0, \textbf{E}_1, \cdots]$. It can be seen that  $\overline{|\bm{\mathcal{E}}[:,0]|}$ increases substantially as $h$ increases. Since we provide all inputs to the decoder’s self-attention, a large input size may cause \textit{attention dilution} \cite{qin_devil_2022}. With more inputs, the attention weights must be distributed across a larger number of inputs, resulting in each input receiving a smaller share of the attention, making it harder for the model to focus on the most relevant inputs.

Next, we combined all effective sources of patient data (those  providing a significant improvement). This included `triage', `reconciled medicines', `indication', and `history'. The best performance was observed with no prior exams ($h=0$), indicating that using any prior exams in combination with other sources is detrimental with our model, possibly due to attention dilution. With $h=0$, the combination of all effective sources outperformed each individual source. We then conducted an ablation study using `Images + effective sources ($h=0$)’, which demonstrated that removing any individual patient data source did not result in a significant change in performance.

Following this, we further trained `Images + effective sources (h = 0)' with reinforcement learning (RL), as described in Subsection \ref{sec:body_experiment_setup}. Its performance is shown in Table \ref{tab:comparison}; a CXR-BERT and BERTScore composite reward was used, which demonstrates a marked improvement for each metric, except ARN. The low ARN indicates that this reward introduced repetitions. We also propose to calculate the reward separately for the findings and impression section, as described in Appendix \ref{sec:rewards}. While this produces similar results for the findings section, as shown in Table \ref{tab:comparison}, this significantly improves the scores on the impression section section, as shown in Table \ref{tab:rl_impression}. Finally, we incorporate ARN into the composite reward. This effectively reduces repetitions, as evidenced by the improved ARN, albeit with a slight trade-off in the other metrics. Compared to other benchmark CXR report generation models in the literature that included MIMIC-CXR in their training data, our model significantly outperformed them on multiple metrics in Table \ref{tab:comparison}, despite having substantially fewer training samples. This demonstrates the impact of incorporating auxiliary patient data on CXR report generation.

Figure \ref{fig:case_study} demonstrates how auxiliary patient data enhances CXR report generation. Mild pulmonary edema was identified only when this data was incorporated. The patient’s low oxygen saturation, chief complaint of congestive heart failure (CHF) --- a common cause of pulmonary edema --- and reconciled medicines (\textit{Furosemide}, \textit{Metoprolol Tartrate}, \textit{Lisinopril}, \textit{Spironolactone}) indicate active management of fluid overload and cardiac dysfunction, all pointing to pulmonary edema. This supplementary evidence allowed the model to corroborate the imaging findings and  identify pulmonary edema.

In Appendix \ref{sec:error_analysis}, we perform an error analysis to assess the influence of auxiliary patient data on the generated reports. Our findings show that incorporating auxiliary patient data increases the AUC for 10 out of the 14 CheXpert labels (Figure \ref{fig:chexpert_score_diff}), demonstrating its utility across multiple pathologies. Additionally, we analysed its impact on the generated reports for eight exams, with the following key observations:\\
\textbf{True positives ($n=2$):} The model utilised supportive auxiliary patient data effectively. (See Appendix \ref{sec:tp_eg1} and \ref{sec:tp_eg2}.)\\
\textbf{False positives ($n=2$):} The model was misled by confounding auxiliary patient data. (See Appendix \ref{sec:fp_eg1} and \ref{sec:fp_eg2}.)\\
\textbf{True negatives ($n=2$):} The model correctly ignored confounding auxiliary patient data. (See Appendix \ref{sec:tn_eg1} and \ref{sec:tn_eg2}.)\\
\textbf{False negatives ($n=2$):} The model failed to leverage supportive auxiliary patient data. (See Appendix \ref{sec:fn_eg1} and \ref{sec:fn_eg2}.)\\
Auxiliary patient data sources—including the indication and history sections, triage data, and reconciled medicines—collectively contributed to the model’s predictions. No single source consistently dominated in providing evidence, with the interplay between these sources frequently complementing one another. A critical challenge for the model lies in its ability to appropriately balance the auxiliary patient data evidence with imaging evidence, particularly when conflicting signals are present. To address this limitation, we propose two key improvements: increasing the size of the training dataset, which is currently relatively small, and adopting an LLM-based decoder. LLMs offer advanced reasoning capabilities, enabling them to better synthesise and prioritise evidence from diverse sources. 

Table \ref{tab:triage} compares different methods for converting value and category columns into embeddings using the triage and reconciled medicines table, as these contain multiple value and category columns. The aforementioned method of producing embeddings by grouping data from value and category columns (`Grouped embeddings') is compared to two other methods. The first is separate embeddings for each datum, where each value column datum is separately transformed using the previously described FNN, while each category column datum is converted to an embedding using a learnable weight matrix, akin to how token embeddings are produced (`Separate embeddings'). The second method modifies `Separate embeddings' by instead converting the value column data to text and using the decoder's tokenizer and token embeddings (`Values-to-text, categories-to-tokens'). The results indicate that the grouped embeddings method was the best representation of heterogeneous patient data for a multimodal language model.

% \begin{table}
%     \centering
%     \begin{tabular}{cp{2in}}
        
%     \end{tabular}
%     \caption{Caption}
%     \label{tab:my_label}
% \end{table}

\section{Conclusion} 

This paper demonstrates the value of incorporating diverse patient data into automated CXR report generation. 
By integrating patient data from the MIMIC-CXR and MIMIC-IV-ED datasets, we have shown significant improvements in the diagnostic accuracy of generated radiology reports. 
Our empirical evaluation uncovers new sources of patient information that enhance CXR report generation, including data from ED stays, triaging information, aperiodic vital signs, medicines, and the history section of radiology reports.
We present specific methods to convert multimodal patient data into embeddings for a language model, encompassing numerical, categorical, textual, temporal, and image data.
We encourage further research and experimentation using our released dataset splits, code, and model checkpoints to explore innovative methods for multimodal patient data integration, with the ultimate goal of enhancing diagnostic accuracy and patient care.

\begin{table}[t]
\caption{\label{tab:triage}Patient data embedding strategies.
% Four training runs were used ($n=3\,832; ~{\rm exams} ~958 \times 4~{\rm runs}$). 
\underline{Underlined} indicates a stat. sig. difference to `Baseline' ($p<0.05$).}
\vspace{-10pt}
\small
\begin{tabular}{@{}p{1.3in}llll@{}}
\toprule
Embeddings & RG & CX  & CB & BS \\ \midrule

\multicolumn{5}{c}{\cellcolor[RGB]{200,200,200} \textit{Images}} \\
Baseline & 29.00 & 25.81  & 59.04 & 23.85\\
\multicolumn{5}{c}{\cellcolor[RGB]{200,200,200} \textit{Images + triage + reconciled medicines}} \\
Grouped embeddings & \textbf{31.69} & \textbf{\underline{26.72}}  & \textbf{\underline{64.01}} & 24.38\\
\midrule
Separate embeddings & 25.28 & 25.32  & \underline{46.29} & 23.51\\
\midrule
Values-to-text, categories-to-embeddings & 30.70 & \underline{26.46}  & 58.62 & \textbf{\underline{24.58}}\\

\bottomrule

\end{tabular}
\end{table}

\clearpage

\section{Limitations}

Despite the promising results demonstrated in this study, several limitations must be acknowledged. Firstly, the generalisability of our findings may be constrained by the datasets utilised, specifically MIMIC-CXR and MIMIC-IV-ED, which are derived from a single institution, the Beth Israel Deaconess Medical Center. This could introduce biases unique to the demographic and clinical practices of this institution, potentially limiting the applicability of our model to other healthcare settings with different patient populations or clinical workflows. Our reliance on these datasets is due to the fact that they are the only publicly available sources that link CXR exams with ED records.

This study currently lacks subjective evaluation by radiologists, which is essential for assessing the quality of generated reports. We plan to address this by evaluating with a private dataset and conducting radiologist-led assessments. To facilitate this, we are securing agreements and ethics approval for access to patient data and radiologist time. However, this process is extensive and beyond the scope of this study, and will instead be used to subjectively evaluate future models.

Another limitation pertains to the completeness and quality of the patient data. Despite incorporating a wide range of data sources, the datasets still contain missing or incomplete information, which can affect model performance. For example, not all exams include a history section, and not all ED patient records have administered medicines available, leading to potential gaps in the data that the model can utilise. However, this reflects the nature of real patient records where issues of data quality and completeness are to be expected.

Our model’s architecture, while effective, has certain limitations. It struggles with large input sizes, especially when incorporating multiple prior exams, likely due to attention dilution. It also at times struggles with supportive or confounding evidence from the auxiliary patient data, introducing false positive or false negative predictions. Future work should explore advanced attention mechanisms, hierarchical models, and LLMs to better manage large input sequences and to better balance auxiliary patient data evidence with imaging evidence.

The interpretability of the model also poses a challenge. While our model shows improved diagnostic accuracy, the decision-making process within the multimodal language model remains a black box. Developing methods to enhance the interpretability and explainability of the model's outputs would be beneficial, especially in clinical settings where understanding the rationale behind a diagnosis is critical.

Finally, while we provide a comprehensive set of metrics to evaluate our model's performance, these metrics focus primarily on the diagnostic accuracy and quality of the generated reports. Broader evaluations considering clinical outcomes, such as the impact on patient management or reduction in radiologist workload, would offer a more holistic view of the benefits and limitations of CXR report generation models in general. Conducting such assessments could help to better understand the practical implications of deploying these models in a clinical setting.

In summary, while our study provides valuable insights into the integration of multimodal patient data for CXR report generation, addressing these limitations will be crucial for further advancements and broader adoption of such models in clinical practice. Future research should explore alternative architectures and training strategies, find alternative datasets to evaluate generalisability, improve model interpretability, and comprehensively assess the practical impact on patient care and radiologist workflow.

\section{Ethical Considerations}
In this research, we used real-world patient data from the MIMIC-CXR and MIMIC-IV-ED datasets. Since these datasets are de-identified, we consider privacy leakage risks to be minimal. Our method employs a language model to generate medical reports from patient data. However, we acknowledge that language models can exhibit bias and produce hallucinations, which may result in incorrect content in the generated reports.

% \subsection{Appendices}

% Use 

\bibliography{bibliography}

\appendix
\counterwithin{figure}{section}
\counterwithin{table}{section}

\section{Dataset Details
} \label{sec:dataset_details}
Each of the exams for the dataset described in Section \ref{sec:dataset} had one ED stay and triage row; 53\% had at least one reconciled medicines row with up to 106 rows; 62\% had at least one vital signs row with up to 69 rows; and 37\% had at least one administered medicines row with up to 52 rows. Exams had an indication section 66\% of the time with a maximum of 75 words, a history section 34\% of the time with a maximum of 74 words, and a comparison section 97\% of the time with a maximum of 129 words. Only one exam had both an indication and a history section.

\section{Prior Exam Embeddings} \label{sec:appendix_embeddings}
The images, findings section, and impression section from previous exams were considered. For prior exams, the time delta was positive, calculated by subtracting the time of the prior exam from the current exam. The images, findings section, and impression section from prior exams were given distinct source embeddings, separate from the current exam, to enhance differentiation. The comparison section from the current exam was also investigated, anticipating that references to prior exams in this section would prompt the decoder to reflect this in the generated report. We explored prior exams with a history size \( h \) of up to three. Note that all exams from MIMIC-CXR were considered for the priors (train/validation/test $222\,758$/$1\,808$/$3\,269$ exams), including those that did not occur during an ED stay and those that did not have a findings and/or impression section. 

\section{Table Column Determination} \label{sec:appendix_columns}

The columns from the tables described in Figure \ref{fig:data} were given the following designations:

\begin{itemize}
    \item For the ED stay table, the patients `intime' was used as the event time. Gender (e.g., `F'), race (e.g., `HISPANIC OR LATINO'), and arrival transport (e.g., `AMBULANCE') were designated as category columns. The disposition column was not considered. 
    \item For the triage table, the `intime' from the ED stay table was used. Temperature (e.g., `100.6'), heart rate (e.g., `93'), respiratory rate (e.g., `16'), O2 saturation (e.g., `94'), systolic blood pressure (SBP) (e.g., `110'), diastolic blood pressure (DBP) (e.g., `56'), and acuity (e.g., `2') were designated as value columns. Pain (e.g., `6-9' and `yes.') and the chief complaint (e.g., `BILATERAL FOOT PAIN') were designated as text columns. 
    \item The column designations for the aperiodic vital signs table were identical to the triage table, except for the rhythm column (e.g., `Normal Sinus Rhythm'), which was treated as a category column. The aperiodic vital signs table also had no chief complaint column and the `charttime' column was used as the event time.
    \item For the reconciled medicines table, the `intime' from the ED stay table was used as the event time, as it pertains to the patient's medicine history prior to the ED stay. The name column was designated as a text column, while the gsn, ndc, etc\_rn, and etccode columns were designated as category columns. The etcdescription column was not considered, as it is a descriprion of the etccode column.
    \item For the administered medicines (\texttt{pyxis}) table, `charttime' was used as the event time. The med\_rn, name, gsn\_rn, and gsn columns were all treated as category columns. The name column for the administered medicines column did not have as high of a cardinality as the name column from the reconciled medicines column, allowing it to be considered as a category column.
    \item For the metadata table, the `PerformedProcedureStepDescription', `ViewPosition', `ProcedureCodeSequence\_CodeMeaning', `ViewCodeSequence\_CodeMeaning', and `PatientOrientationCodeSequence\_CodeMeaning' columns were considered, and designated as category columns.
\end{itemize}

\section{Experiment Setup} \label{sec:appendix_experiment_setup}

\subsection{Metrics} 

GREEN \cite{ostmeier-etal-2024-green}, CheXbert-F1 \citep{smit_combining_2020}, RadGraph-F1 \cite{delbrouck_improving_2022}, BLEU-4 \citep{papineni_bleu_2001}, and BERTScore-F1 (\texttt{roberta-large\_L17\_no-idf\_rescaled}) \citep{zhang_bertscore_2019} have been found to correlate with radiologists’ assessment of reporting \citep{Yu2022.08.30.22279318, ostmeier-etal-2024-green} and were a part of our evaluation. Additionally, we include CXR-BERT \cite{boecking_making_2022, nicolson_longitudinal_2024}, and ROUGE-L \citep{lin_automatic_2003}. GREEN, CheXbert-F1, RadGraph-F1, and CXR-BERT were intended to capture the clinical semantic similarity between the generated and radiologist reports, while BERTscore-F1 was intended to capture general semantic similarity. Finally, ROUGE-L and BLEU-4 were intended to capture the syntactic similarity between the generated and radiologist reports. We also propose a new metric that measures \textit{n}-gram repetition rate, namely the absence of repeated \textit{n}-grams (ARN). It is calculated as:
\begin{equation}
\text{ARN} = 
\begin{cases} 
1.0 & \text{if } L < n, \\
1.0 - \frac{\sum_{i=1}^{M} (\text{Count}(g_i) - 1)}{M} & \text{if } L \geq n,
\end{cases}
\end{equation}
where $L$ is the total number of tokens in the generated report, $n$ is the  $n$-gram size, $M = L - n + 1$ is the total number of $n$-grams in the report, $g_i$ is the $i^{th}$ unique $n$-gram in the report, $\textrm{Count}(g_i)$ is the $n$-gram frequency in the report. The tokenizer described in Appendix \ref{sec:appendix_model} was used with an $n$-gram size of three.

For the models in Table \ref{tab:comparison} that generate a report for each image in an exam, the average score was taken across all reports for an exam. Following this, the final average score was computed across all exams for both models that generate a report per image and those that generate a report per exam.

For CheXbert, the macro-averaged F1 was computed between the 14 CheXbert observations extracted from the generated and radiologist reports. ``No mention'', ``negative'', and ``uncertain'' were considered negative, while ``positive'' was considered positive. Here, the true positives, false positives, and false negatives were averaged over the reports of each exam for the models that generate a report per image. 

We also perform statistical testing; first, a Levene’s test was conducted to reveal if the variances across model scores was homogeneous or not. If the assumption of equal variances was upheld, a one-way ANOVA was conducted to determine if there was a significant difference between models. Finally, pairwise Tukey-HSD post-hoc tests were used for pairwise testing. If the assumption of equal variances was violated, a one-way Welch’s ANOVA was conducted to determine if there was a significant difference between models. Finally, Games-Howell post hoc tests were used for pairwise testing. A $p$-value of 0.05 was used for all significance testing. Statistical testing was not performed for CheXbert, as it is a classification metric.

\subsection{Model} \label{sec:appendix_model}

Our model is illustrated in Figure \ref{fig:model}; following \citet{nicolson_rrg24_2024}, we utilised UniFormer as the image encoder (in particular, the $384 \times 384$ base model warm started with its token labelling fine-tuned checkpoint) \cite{li_uniformer_2023}. The image embeddings are formed by processing each image in the exam separately with the image encoder and then projecting its last hidden state to match the decoder's hidden size using a learnable weight matrix. Each image was resized using bicubic interpolation so that its smallest side had a length of 384 and its largest side maintained the aspect ratio. Next, the resized image was cropped to a size of $\mathbb{R}^{3 \times 384 \times 384}$. The crop location was random during training and centred during testing. Following \cite{elgendi_effectiveness_2021}, the image was rotated around its centre during training, where the angle of rotation was sampled from $\mathcal{U}{[{-5^{\circ}, 5^{\circ}}]}$. Finally, the image was standardised using the statistics provided with the UniFormer checkpoint. A maximum of five images per exam were used during training. If more were available, five were randomly sampled uniformly without replacement from the exam for each epoch.

Again following \cite{nicolson_rrg24_2024}, we employed the Llama architecture for the decoder, which is notable for features such as its rotary positional encoding (RoPE), root mean square normalisation (RMSNorm), and SwiGLU activation function \cite{touvron_llama_2023}. A byte-level byte pair encoding tokenizer \citep{Wang_Cho_Gu_2020} was trained with a vocabulary size of $30\,000$. It was trained on the findings, impression, indication, and history sections (not the comparison section) of the entire MIMIC-CXR training set, as well as the `pain' and `chiefcomplaint' columns from the triage table, the `name' column of the reconciled medicines  table, and the `pain' column from the vital signs table (from the entire MIMIC-IV-ED dataset). Newline, tab, repeated whitespaces, and leading and trailing whitespaces were removed from any text before tokenization.

\begin{figure}
    \centering
    \includegraphics[scale=1.0]{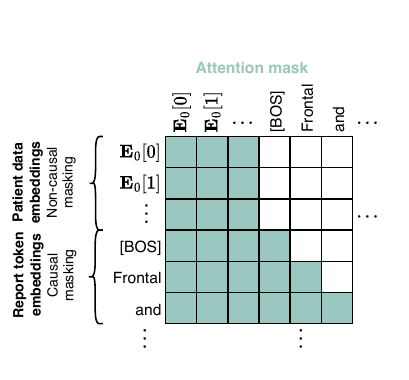}
    \caption{\label{fig:attention_mask}Attention mask for the decoder. Non-causal masking was used for the patient data embeddings and causal masking for the report token embeddings.}
\end{figure}

The hyperparameters of the Llama decoder were six hidden layers, a hidden size of 768, 12 attention heads per layer, and an intermediate size of $3\,072$. The maximum number of position embeddings was set to $2\,048$ to accommodate all the patient data embeddings and the report tokens. The maximum number of tokens that could be generated was set to 256, which was also the limit for the radiologist reports during training. During testing, a beam size of four was utilised. The Llama decoder allows a custom attention mask to be provided in current implementations.\footnote{https://huggingface.co/blog/poedator/4d-masks} This enabled non-causal masking to be utilised for the prompt and causal masking for the report token embeddings, as shown in Figure \ref{fig:attention_mask}. This ensured that the self-attention heads were able to attend to all of the patient data embeddings at each position.

\subsection{Training} 

Three stages of training were performed. Each stage used \textit{AdamW}~\citep{loshchilov_decoupled_2022} for mini-batch gradient descent optimisation and gradient clipping with a maximum norm of 1.0 to prevent exploding gradients and maintain training stability. Training and evaluation was performed on a 94GB NVIDIA H100 GPU. The three stages were as follows:

\begin{enumerate}
    \item Teacher forcing (TF) \cite{williams_learning_1989} was performed on the MIMIC-CXR dataset with only the images for each exam as input, and exams that contained both a findings and impression section. This gave a training/validation split of $232\,855$/$1\,837$ images, $125\,417$/991 exams, and $57\,102$/436 patients. Training was performed with an initial learning rate of 5e-5, a mini-batch size of 8, a maximum of 32 epochs, and with float16 automatic mixed precision. All model parameters were trainable during this stage. The validation macro-averaged CheXbert-F1 was the monitored metric for checkpoint selection. This stage was necessary, as the language model struggled to generate reports from multiple patient data sources without prior learning.
    \item TF was used in the second stage of training, with the MIMIC-CXR \& MIMIC-IV-ED dataset described in Section \ref{sec:dataset} with the inputs described in Table \ref{tab:main_results}. The training strategy was identical to the previous stage, except that a maximum of 16 epochs was performed, and the image encoder's parameters were frozen (except for its projection). The models featured in Table \ref{tab:main_results} were trained using only the first two stages.
    \item Reinforcement learning using self-critical sequence training (SCST) \cite{rennie_self-critical_2017} was performed with the rewards described in Appendix \ref{sec:rewards} in the final stage of training. The sample report for SCST was generated with top-\textit{k} sampling ($k=50$). Training was performed with an initial learning rate of 5e-6, a mini-batch size of 32, a maximum of 32 epochs, and with float32 precision. A warmup phase of $5\,000$ training steps was used for the learning rate, linearly increasing from zero. The image encoder's parameters were frozen during this stage (except for its projection). The validation BERTScore-F1 was the monitored metric for checkpoint selection. This stage of training was only applied to the best model from Table \ref{tab:main_results}, `Images + effective sources (h = 0)', with the results presented in Table \ref{tab:comparison}.
\end{enumerate}

\subsection{Comparison Models} 

The generated reports for the models in Table \ref{tab:comparison} were attained as follows:
\begin{itemize}
    \item EMNLI reports were generated following \url{https://github.com/ysmiura/ifcc} \citep{miura_improving_2021}.
    \item CMN reports were generated following \url{https://github.com/zhjohnchan/R2GenCMN} \citep{chen_cross-modal_2021}.
    \item TranSQ reports were kindly provided by the authors \cite{wang_transq_2022}.
    \item RGRG reports were generated following \url{https://github.com/ttanida/rgrg} \cite{tanida_interactive_2023}.
    \item CvT2DistilGPT2 reports were generated following \url{https://github.com/aehrc/cvt2distilgpt2} \cite{nicolson_improving_2022}.
    \item RaDialog reports were kindly provided by the authors \cite{pellegrini_radialog_2023}.
    \item MedXChat reports were kindly provided by the authors \cite{yang_medxchat_2023}.
    \item CXR-LLaVA-v2 reports were generated following \url{https://huggingface.co/ECOFRI/CXR-LLAVA-v2} \cite{lee_cxr-llava_2024}.
    \item CXRMate reports were generated following \url{https://huggingface.co/aehrc/cxrmate} \cite{nicolson_longitudinal_2024}.
    \item CXRMate-RRG24 reports were generated following \url{https://huggingface.co/aehrc/cxrmate-rrg24} \cite{nicolson_rrg24_2024}.
\end{itemize}
CXRMate-RRG24 was trained on five datasets, including MIMIC-CXR. RGRG was trained on the ImaGenome dataset derived from MIMIC-CXR --- which may have some overlap with our test set.

\section{Reinforcement Learning Rewards} \label{sec:rewards}

The separate reward per section was calculated as:
\begin{equation}
    \begin{split}
r_s(\hat{\textbf{w}}_f,\textbf{w}_f,\hat{\textbf{w}}_i,\textbf{w}_i)=&\alpha_1\cdot r_f(\hat{\textbf{w}}_f,\textbf{w}_f) + \\ &\alpha_2\cdot r_i(\hat{\textbf{w}}_i,\textbf{w}_i),   \\
    \end{split}
\end{equation}
where $r_s(\cdot)$ is the composite reward for the sections of the report, $r_f(\cdot)$ is the reward for the findings section, and $r_i(\cdot)$ is the reward for the impression section, $\hat{\textbf{w}}_f$ is the generated findings section, $\textbf{w}_f$ is the radiologist findings section, $\hat{\textbf{w}}_i$ is the generated impression section, $\textbf{w}_i$ is the radiologist impression section, and $\alpha_1$ and $\alpha_2$ are weights. Normally, $r_r(\hat{\textbf{w}}_r,\textbf{w}_r)$ is calculated, where $\hat{\textbf{w}}_r$ and $\textbf{w}_r$ are the generated and radiologist reports, which include both the findings and impression sections. 

The reward $r_f(\cdot)$, $r_i(\cdot)$, or $r_r(\cdot)$ is calculated as:
\begin{equation}
    \begin{split}
    r(\hat{\textbf{w}},\textbf{w}) = &\lambda_1\cdot\textrm{CXR-BERT}(\hat{\textbf{w}},\textbf{w})+ \\
    &\lambda_2\cdot\textrm{BERTScore}(\hat{\textbf{w}},\textbf{w})+ \\
    &\lambda_3\cdot\textrm{ARN}(\hat{\textbf{w}},\textbf{w}), \\
    \end{split}
\end{equation}
where $\lambda_1$, $\lambda_2$, and $\lambda_3$ are weights. For `Images + effective source ($h=0$) + RL with CXR-BERT + BERTScore reward', $\lambda_1=0.5$, $\lambda_2=0.5$, and $\lambda_3=0.0$. For `Images + effective source ($h=0$) + RL with CXR-BERT + BERTScore reward per section', $\alpha_1=0.75$, $\alpha_2=0.25$, $\lambda_1=0.5$, $\lambda_2=0.5$, and $\lambda_3=0.0$. A higher weight was used for the findings section, as it is longer on average than the impression section. For `Images + effective source ($h=0$) + RL with CXR-BERT + BERTScore + ARN reward per section', $\alpha_1=0.75$, $\alpha_2=0.25$, $\lambda_1=0.45$, $\lambda_2=0.45$, and $\lambda_3=0.1$. Only a weak contribution of the ARN was required to prevent repetitions.

The improvement that separating the reward per section has on the findings section is negligible, as seen in Table \ref{tab:comparison}. However, separating the reward per section improves the scores for the impression section, as shown in Table \ref{tab:rl_impression}. Separating the reward likely enables the model to better optimise for the concise and summarised nature of the impression section, which was previously overshadowed by the dominance of the findings section’s requirement for comprehensive detail when both were jointly considered.

\begin{table*}[t]
\centering
 \vspace{-5pt}
\caption{\label{tab:rl_impression}Impact of the reward on the impression section of the test set described in Section \ref{sec:dataset} ($n=9\,580 ~{\rm exams}; ~958 \times 10~{\rm runs}$ for `Images + effective sources ($h=0$)', $n=1\,916 ~{\rm exams}; ~958 \times 3~{\rm runs}$ for the remaining models). Evaluation is on the \textbf{impression} section only.}
\small
\setlength{\tabcolsep}{5pt} % Reduce column spacing

\begin{tabular}{@{}lllllllll@{}}
\toprule
 Model & RG & CX & CB & G & BS & R-L & B4 & ARN \\ \midrule
 % Model & CX & CB & BS & R-L & B4 & ARN \\ \midrule

Images + effective sources ($h=0$) & \cellcolor[RGB]{242,240,237} 20.21 & \cellcolor[RGB]{239,240,242} 26.81 & \cellcolor[RGB]{242,240,237} 57.61 & \cellcolor[RGB]{239,240,242} 28.71 & \cellcolor[RGB]{242,240,237} 27.90 & \cellcolor[RGB]{239,240,242} 25.02 & \cellcolor[RGB]{242,240,237} 4.77 & \cellcolor[RGB]{190,215,237} 99.59\\
\quad + RL (CXR-BERT + BERTScore reward) & \cellcolor[RGB]{250,210,91} 23.96 & \cellcolor[RGB]{229,235,241} 28.07 & \cellcolor[RGB]{244,233,203} 62.85 & \cellcolor[RGB]{171,205,235} 30.58 & \cellcolor[RGB]{245,231,193} 31.58 & \cellcolor[RGB]{152,195,233} 28.48 & \cellcolor[RGB]{255,194,10} \textbf{7.84} & \cellcolor[RGB]{12,123,220} \textbf{99.89}\\
\quad\quad + reward per section & \cellcolor[RGB]{255,194,10} \textbf{24.89} & \cellcolor[RGB]{127,182,231} 31.08 & \cellcolor[RGB]{255,194,10} \textbf{71.12} & \cellcolor[RGB]{147,193,233} 30.89 & \cellcolor[RGB]{254,194,11} 36.27 & \cellcolor[RGB]{38,136,222} 30.27 & \cellcolor[RGB]{247,221,147} 6.70 & \cellcolor[RGB]{239,240,242} 99.33\\
\quad\quad\quad + ARN reward & \cellcolor[RGB]{254,194,11} 24.87 & \cellcolor[RGB]{12,123,220} \textbf{32.88} & \cellcolor[RGB]{255,194,10} \textbf{71.12} & \cellcolor[RGB]{12,123,220} \textbf{32.14} & \cellcolor[RGB]{255,194,10} \textbf{36.31} & \cellcolor[RGB]{12,123,220} \textbf{30.61} & \cellcolor[RGB]{248,219,133} 6.84 & \cellcolor[RGB]{57,146,224} 99.83\\

% Images + effective sources ($h=0$) & \cellcolor[RGB]{242,240,237} 26.81 & \cellcolor[RGB]{239,240,242} 57.61 & \cellcolor[RGB]{242,240,237} 27.90 & \cellcolor[RGB]{239,240,242} 25.02 & \cellcolor[RGB]{242,240,237} 4.77 & \cellcolor[RGB]{0,0,0} nan\\
% \quad + RL (CXR-BERT + BERTScore reward) & \cellcolor[RGB]{243,238,227} 28.07 & \cellcolor[RGB]{205,223,239} 62.85 & \cellcolor[RGB]{245,231,193} 31.58 & \cellcolor[RGB]{152,195,233} 28.48 & \cellcolor[RGB]{255,194,10} \textbf{7.84} & \cellcolor[RGB]{12,123,220} \textbf{99.89}\\
% \quad\quad + reward per section & \cellcolor[RGB]{248,217,124} 31.08 & \cellcolor[RGB]{12,123,220} \textbf{71.12} & \cellcolor[RGB]{254,194,11} 36.27 & \cellcolor[RGB]{38,136,222} 30.27 & \cellcolor[RGB]{247,221,147} 6.70 & \cellcolor[RGB]{239,240,242} 99.33\\
% \quad\quad\quad + ARN reward & \cellcolor[RGB]{255,194,10} \textbf{32.88} & \cellcolor[RGB]{12,123,220} \textbf{71.12} & \cellcolor[RGB]{255,194,10} \textbf{36.31} & \cellcolor[RGB]{12,123,220} \textbf{30.61} & \cellcolor[RGB]{248,219,133} 6.84 & \cellcolor[RGB]{57,146,224} 99.83\\

\bottomrule
 
\end{tabular}
\end{table*}

\begin{figure*}[t]
    \centering
    \includegraphics[scale=1.0]{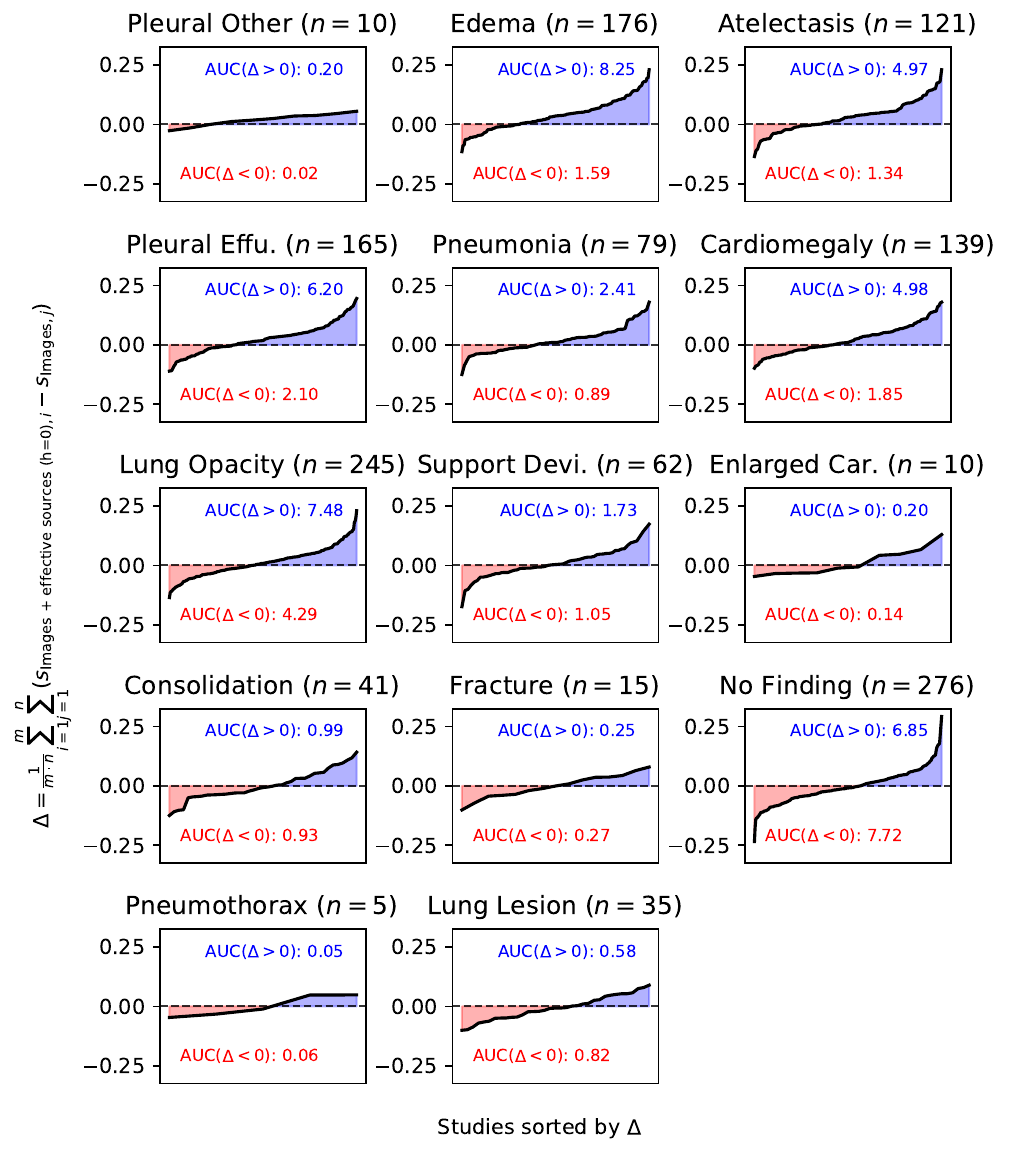}
    \caption{\label{fig:chexpert_score_diff} The mean pairwise difference GREEN score for the generated report (findings and impression sections) of each exam from the test set between 10 training runs of the ``Images” model and the ``Images + effective sources (h=0)” model. This illustrates the performance change (increase or decrease) over the exams resulting from incorporating auxiliary patient data for different CheXpert labels. $\Delta$, $m$ and $n$ are the number of training runs for each model ($m=n=10$) and $s$ is the GREEN score for one of the models. The subplots are sorted in descending order based on the ratio of $\textrm{AUC}(\Delta>0)$ to $\textrm{AUC}(\Delta<0)$.}
\end{figure*}

\begin{figure*}[t]
    \centering
    \includegraphics[scale=0.75]{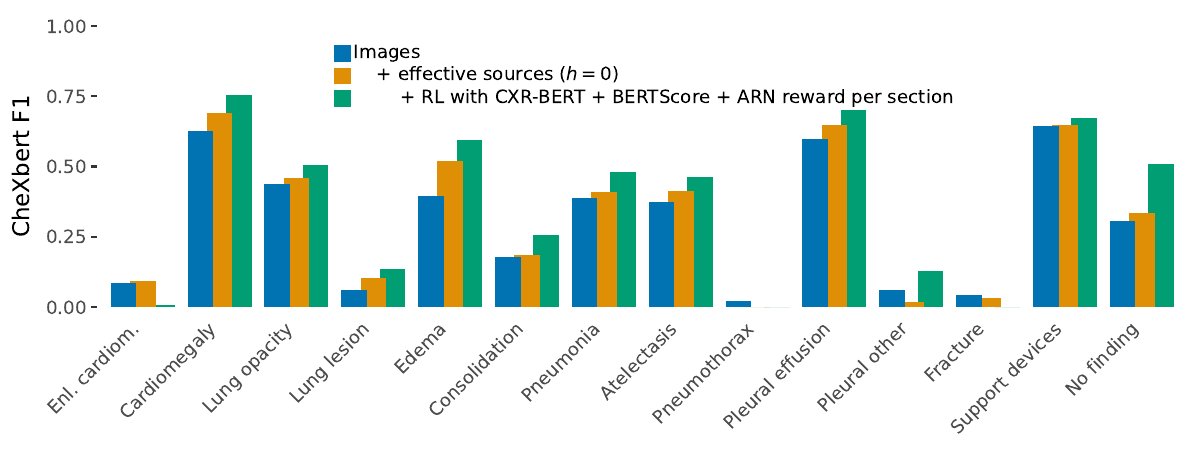}
    \caption{\label{fig:chexbert_labels}F1-score for each CheXbert label. ($n=9\,580 ~{\rm exams}; ~958 \times 10~{\rm runs}$ for `Images' and `Images + effective sources ($h=0$)' and $n=2\,874 ~{\rm exams}; ~958 \times 3~{\rm runs}$ for `Images + effective sources ($h=0$) + RL with CXR-BERT + BERTScore + ARN reward per section'.}
\end{figure*}

\section{Ancillary Results}

In Figure \ref{fig:chexbert_labels}, the F1-scores for each CheXbert label are shown. The `Images + effective sources ($h=0$)' model from Table \ref{tab:main_results} attained a higher score than the `Images’ model for 11 of the 14 labels. This suggests that incorporating auxiliary patient data from MIMIC-IV-ED and MIMIC-CXR provides a general improvement, rather than benefiting any specific pathology.

Further improvements can be seen for most labels when reinforcement learning (RL) is used (i.e., our model from Table \ref{tab:comparison}). However, there are performance decreases for `enlarged cardiomediastinum’, `pneumothorax’, and `fracture’. This might be due to these pathologies being underrepresented in the MIMIC-CXR dataset, leading the model to optimise for more common pathologies during reinforcement learning.

The results for exams that include an aperiodic vital signs table are show in Table \ref{tab:vitalsign}. Adding it produced a significant improvement in the scores for CXR-BERT, indicating that it should be considered if available. The results for exams that include an administered medicines table are show in Table \ref{tab:pyxis}. Adding did not produce a significant improvement in the scores, indicating that it is not useful for CXR report generation.

\begin{table*}[t]
\centering
\caption{\label{tab:vitalsign}Results for exams that have an aperiodic vital sign table ($n=5\,250; ~{\rm studies} ~525 \times 10~{\rm runs}$). \underline{Underlined} scores indicate a significant difference to the scores of `Images' ($p<0.05$).}
\small
\begin{tabular}{@{}llllllll@{}}
\toprule

 Model & RG & CX & CB & G & BS & R-L & B4 \\ \midrule

Images & \cellcolor[RGB]{255,194,10} \textbf{24.73} & \cellcolor[RGB]{239,240,242} 29.41 & \cellcolor[RGB]{242,240,237} 58.63 & \cellcolor[RGB]{239,240,242} 35.11 & \cellcolor[RGB]{255,194,10} \textbf{24.33} & \cellcolor[RGB]{239,240,242} 25.85 & \cellcolor[RGB]{255,194,10} \textbf{4.89}\\
Images + vital signs & \cellcolor[RGB]{242,240,237} 24.55 & \cellcolor[RGB]{12,123,220} \textbf{29.73} & \cellcolor[RGB]{255,194,10} \textbf{\underline{60.32}} & \cellcolor[RGB]{12,123,220} \textbf{35.21} & \cellcolor[RGB]{242,240,237} 24.17 & \cellcolor[RGB]{12,123,220} \textbf{25.97} & \cellcolor[RGB]{242,240,237} 4.87\\

\bottomrule
 
\end{tabular}
\end{table*}

\begin{table*}[t]
\centering
\caption{\label{tab:pyxis}Results for exams that have a administered medicines table ($n=3\,520; ~{\rm studies} ~352 \times 10~{\rm runs}$). \underline{Underlined} scores indicate a significant difference to the scores of `Images' ($p<0.05$).}
\small
\begin{tabular}{@{}llllllll@{}}
\toprule

 Model & RG & CX & CB & G & BS & R-L & B4 \\ \midrule

Images & \cellcolor[RGB]{255,194,10} \textbf{25.19} & \cellcolor[RGB]{239,240,242} 28.29 & \cellcolor[RGB]{242,240,237} 59.24 & \cellcolor[RGB]{12,123,220} \textbf{36.13} & \cellcolor[RGB]{255,194,10} \textbf{24.81} & \cellcolor[RGB]{12,123,220} \textbf{26.61} & \cellcolor[RGB]{255,194,10} \textbf{5.15}\\
Images + administered medicines & \cellcolor[RGB]{242,240,237} 24.70 & \cellcolor[RGB]{12,123,220} \textbf{29.53} & \cellcolor[RGB]{255,194,10} \textbf{59.53} & \cellcolor[RGB]{239,240,242} 35.82 & \cellcolor[RGB]{242,240,237} 24.46 & \cellcolor[RGB]{239,240,242} 26.38 & \cellcolor[RGB]{242,240,237} 4.85\\

\bottomrule
 
\end{tabular}
\end{table*}

\section{Error analysis} \label{sec:error_analysis}

\subsection{Impact of Auxiliary Patient Data on the CheXpert Labels}

Figure \ref{fig:chexpert_score_diff} demonstrates the impact of incorporating auxiliary patient data for different CheXpert labels. Here, the GREEN score for the `Images + effective sources (h=0)' model is compared to the `Images' model from Table \ref{tab:main_results} for each exam. Note that the generated and radiologist report for each exam will often include findings other than the CheXpert label. Hence, the GREEN scores do not exclusively represent a particular CheXpert label, rather, they represent exams with that label present. The horizontal dashed line where $\Delta = 0$ divides exams where auxiliary patient data improved performance from those where it decreased performance. CheXpert labels with a higher area under the curve (AUC) above the horizontal dashed line suggest that there is a stronger overall benefit from leveraging auxiliary patient data.

Leveraging auxiliary patient data yielded a higher AUC for 10 out of the 14 CheXpert labels, indicating that it is beneficial for many pathologies. For certain CheXpert labels, the influence of auxiliary patient data is less clear, particularly for those associated with smaller sample sizes, such as \textit{enlarged cardiomediastinum} ($n=10$), \textit{consolidation} ($n=10$), \textit{fracture} ($n=15$), \textit{pneumothorax} ($n=5$), and \textit{lung lesion} ($n=35$). The \textit{no findings} AUC of 6.85 for $\Delta>0$ being lower than the AUC of 7.72 for $\Delta<0$ suggests that the auxiliary patient data increases the false positive rate for this model.

\subsection{Impact of Auxiliary Patient Data on the Generated Reports}

To gain a better understanding of how the auxiliary patient data impacts the generated reports, we analyse multiple case studies where it contributes to either true positive, false positive, true negative, or false negative findings in the generated report:
\begin{itemize}
    \item A true positive is where the model has identified a positive occurrence of a pathology that is also identified as positive in the radiologist's report.
    \item A false positive is where the model has incorrectly identified a positive occurrence of a pathology that is not identified as positive in the radiologist's report.
    \item A true negative occurs when a pathology is omitted or absent in the radiologist's report and this is correctly reflected in the generated report, either implicitly through omission or explicitly by stating its absence.
    \item A false negative is where a pathology is positively identified in the radiologist's report but is not positively identified in the generated report.
\end{itemize}
Exams with a high $\Delta$ from Figure \ref{fig:chexpert_score_diff} were selected for true positive and true negative examples, while those with a low $\Delta$ were chosen for false positive and false negative examples.\footnote{Out of the 10 training runs, the `Images + effective sources ($h=0$)' and `Images' models that attained the highest average GREEN score over the test set were selected for the error analysis.} This analysis, though based on only eight exams, exemplifies how auxiliary patient data can both enhance and hinder the CXR report generation process, providing valuable insights into its impact. A more comprehensive analysis would be required to fully characterise the influence of auxiliary patient data across diverse exams and pathologies.

\subsubsection{True Positive: Example 1}\label{sec:tp_eg1}
Table \ref{tab:51707133} demonstrates how auxiliary patient data contributed to the true positive detection of increased interstitial markings, which are suggestive of pulmonary fibrosis. The model not using auxiliary patient data failed to detect the interstitial markings. The patient’s triage data included a respiratory rate consistent with tachypnoea and a chief complaint of dyspnoea, both common features of pulmonary fibrosis. Additionally, the patient’s history of pulmonary fibrosis and worsening shortness of breath provided further context supporting the observed increase in interstitial markings. In this case, the inclusion of auxiliary patient data facilitated a true positive detection.

\begin{table*}[t]
\centering
\caption{\label{tab:51707133} True positive example for study \texttt{51707133}. Here, the triage data and the history section provide additional evidence supporting increased interstitial markings. Only the patient data that \textit{Images + effective sources (h=0)} utilises is shown.}
\small
\setlength{\tabcolsep}{5pt} % Reduce column spacing
\begin{tabular}{p{0.5in}p{5.6in}}
\toprule
\multicolumn{2}{c}{\cellcolor[RGB]{200,200,200} \textit{Patient data}} \\
Image & \includegraphics[scale=0.4, valign=c]{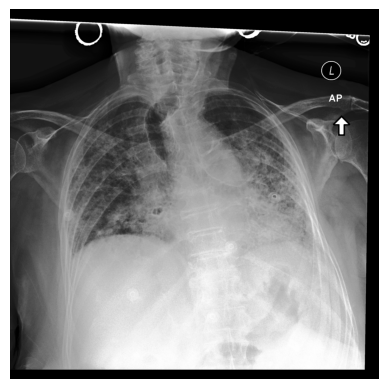} \includegraphics[scale=0.4, valign=c]{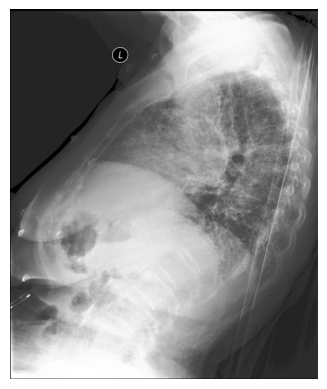} \\\midrule
History & \_\_\_-year-old female with pulmonary fibrosis and CHF with worsening shortness of breath. \\\midrule
Reconciled medicines; name & atorvastatin, azelastine [Astelin], aspirin, calcium carbonate-vitamin D3 [Calcium 500 + D], loratadine, metoprolol succinate, multivitamin, glucosamine sulfate [Glucosamine], acetaminophen, ferrous sulfate [Feosol], torsemide, pantoprazole, lidocaine, ketotifen fumarate. \\\midrule
Triage & 
\begin{tabular}{|l|l|l|l|l|l|l|l|l|}
\centering
temperature &heartrate &resprate &o2sat &sbp &dbp &pain &acuity &chiefcomplaint \\\midrule
99.7 &90.0 &36.0 &100.0 &118.0 &70.0 &0 &2.0 &Dyspnea \\
\end{tabular}\\
\multicolumn{2}{c}{\cellcolor[RGB]{200,200,200} \textit{Radiologist}} \\
Findings & AP and lateral views of the chest. Low lung volumes are seen compatible with patient's history of fibrosis. Diffusely increased interstitial markings are seen throughout the lungs, but these appear overall slightly worse when compared to prior. Cardiomediastinal silhouette is grossly unchanged. No acute osseous abnormality is detected.\\\midrule
Impression & Findings compatible with pulmonary fibrosis with likely superimposed edema. Please note that infection cannot be excluded and clinical correlation is necessary.\\
\multicolumn{2}{c}{\cellcolor[RGB]{200,200,200} \textit{Images + effective sources ($h=0$)} ($\textrm{GREEN}=0.375
$)} \\
Findings &  Frontal and lateral views of the chest. Low lung volumes are again noted. Increased interstitial markings seen more extensive on the left than on the right. There is no evidence of overt consolidation nor effusion. The cardiomediastinal silhouette is stable. No acute osseous abnormality is identified. Degenerative changes are seen at the shoulders.\\\midrule
Impression & Increased interstitial markings throughout the lungs which could be due to chronic lung disease and possible chronic lung disease. No definite superimposed acute process, although clinical correlation suggested.\\
\multicolumn{2}{c}{\cellcolor[RGB]{200,200,200} \textit{Images} ($\textrm{GREEN}=0.091$)} \\
Findings & Assessment is limited due to patient rotation and patient rotation. Lung volumes are low. Heart size appears mildly enlarged. The aorta appears to be calcified. Perihilar haziness and vascular indistinctness is compatible with mild pulmonary edema. Streaky opacities in the lung bases likely reflect areas of atelectasis. No large pleural effusion or pneumothorax is seen. Multilevel degenerative changes are noted in the thoracic spine.\\\midrule
Impression & Mild pulmonary edema and bibasilar atelectasis.\\
\bottomrule
\end{tabular}
\end{table*}

\subsubsection{True Positive: Example 2} \label{sec:tp_eg2}

Table \ref{tab:52841174} demonstrates how auxiliary patient data contributed to the true positive detection of pulmonary edema, which was not detected by the model that does not use auxiliary patient data. Recorded in the patient’s triage data was a respiratory rate consistent with tachypnoea and a chief complaint of dyspnoea (also documented in the history section), both of which are indicative of pulmonary edema. Additionally, furosemide was listed in the patient's reconciled medicines, which is commonly used to manage pulmonary edema. This example underscores how incorporating auxiliary patient data can enhance true positive detection in CXR report generation.

\begin{table*}[t]
\centering
\caption{\label{tab:52841174} True positive example for study \texttt{52841174}. Here, the triage data and reconciled medicines provide additional evidence indicative of pulmonary edema. Only the patient data that \textit{Images + effective sources (h=0)} utilises is shown.}
\small
\setlength{\tabcolsep}{5pt} % Reduce column spacing
\begin{tabular}{p{0.5in}p{5.6in}}
\toprule
\multicolumn{2}{c}{\cellcolor[RGB]{200,200,200} \textit{Patient data}} \\
Image & \includegraphics[scale=0.4, valign=c]{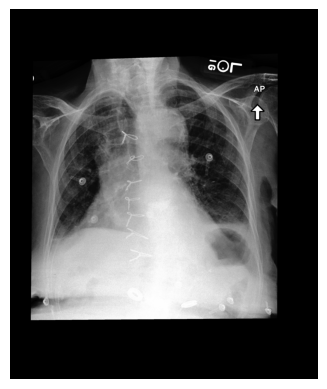} \includegraphics[scale=0.4, valign=c]{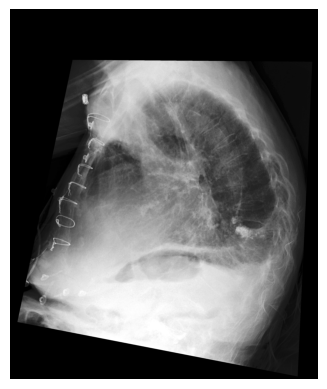} \\\midrule
History & \_\_\_-year-old with dyspnea. \\\midrule
Reconciled medicines; name & Coumadin, furosemide, metoprolol succinate, Calcarb 600 With Vitamin D, simvastatin, Tylenol Extra Strength, levothyroxine, docusate sodium. \\\midrule
Triage & 
\begin{tabular}{|l|l|l|l|l|l|l|l|l|}
\centering
temperature &heartrate &resprate &o2sat &sbp &dbp &pain &acuity &chiefcomplaint \\\midrule
97.0 &81.0 &22.0 &100.0 &102.0 &58.0 &0 &2.0 &DYSPNEA \\
\end{tabular}\\
\multicolumn{2}{c}{\cellcolor[RGB]{200,200,200} \textit{Radiologist}} \\
Findings & AP upright and lateral views of the chest were provided. Midline sternotomy wires are again noted. Patient is rotated somewhat limiting the evaluation of the cardiomediastinal silhouette, though cardiomediastinal silhouette appears grossly stable. There are small layering bilateral effusions with mild interstitial edema. Overall, there has been no significant change from prior study. Bony structures are intact.\\\midrule
Impression & Mild interstitial edema, stable cardiomegaly with small bilateral effusions.\\
\multicolumn{2}{c}{\cellcolor[RGB]{200,200,200} \textit{Images + effective sources ($h=0$)} ($\textrm{GREEN}=0.375$)} \\
Findings & AP upright and lateral views of the chest were provided. Midline sternotomy wires and mediastinal clips as well as a prosthetic cardiac valve. Low lung volumes limit evaluation. There is hilar congestion and mild pulmonary edema. Small bilateral pleural effusions persist. There is left basilar atelectasis. The heart is mildly enlarged. Bony structures appear intact. No free air below the right hemidiaphragm.\\\midrule
Impression & Pulmonary edema, small bilateral pleural effusions, left greater than right.\\
\multicolumn{2}{c}{\cellcolor[RGB]{200,200,200} \textit{Images} ($\textrm{GREEN}=0.222$)} \\
Findings & The patient is status post median sternotomy and CABG. Large hiatal hernia is present. The cardiac silhouette size is mildly enlarged. The aorta is tortuous. Crowding of bronchovascular structures is present with probable mild pulmonary vascular congestion. Small right pleural effusion is present. Patchy opacities in the lung bases may reflect atelectasis. No pneumothorax is demonstrated. There are moderate multilevel degenerative changes seen in the thoracic spine.\\\midrule
Impression & 1. Small right pleural effusion and bibasilar opacities likely reflect atelectasis. Infection at the lung bases cannot be completely excluded. 2. Mild pulmonary vascular congestion. 3. Moderate cardiomegaly.\\
\bottomrule
\end{tabular}
\end{table*}

\subsubsection{False Positive: Example 1} \label{sec:fp_eg1}

Table \ref{tab:51274564} provides an example of where the model leveraging auxiliary patient data introduced a false positive prediction into the generated report. It incorrectly specifies that there are streaky opacities in the lung bases, which are reflective of atelectasis. The model that does not leverage auxiliary patient data did not produce this false positive. Atelectasis typically presents with symptoms such as dyspnoea, tachypnoea, wheezing, and coughing; however, these symptoms were absent from the indication section or the triage data. Although codeine, listed among the patient’s reconciled medicines, can contribute to atelectasis in high doses, there was no evidence of overdose or misuse in this case. This example suggests that weak or ambiguous evidence in the auxiliary data may have influenced the false positive prediction. Further refinement is needed to improve the model’s ability to appropriately weigh auxiliary patient data evidence against imaging evidence.

\begin{table*}[t]
\centering
\caption{\label{tab:51274564} False positive example for study \texttt{51274564}. This example demonstrates how weak auxiliary patient data evidence may have misled the model. Only the patient data that \textit{Images + effective sources (h=0)} utilises is shown.}
\small
\setlength{\tabcolsep}{5pt} % Reduce column spacing
\begin{tabular}{p{0.5in}p{5.6in}}
\toprule
\multicolumn{2}{c}{\cellcolor[RGB]{200,200,200} \textit{Patient data}} \\
Image & \includegraphics[scale=0.4, valign=c]{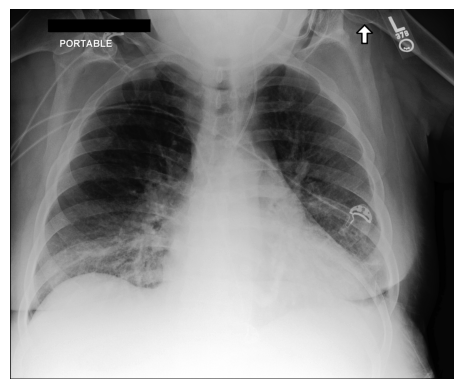} \\\midrule
Indication & Status post new central line placement. \\\midrule
Reconciled medicines; name & colchicine, Aspirin, nifedipine, blood sugar diagnostic [OneTouch Ultra Test], labetalol, calcitriol, insulin needles (disposable) [BD Insulin Pen Needle UF Mini], fluticasone, codeine-guaifenesin, lisinopril, insulin lispro [Humalog KwikPen], insulin glargine [Lantus Solostar], prednisone, acetaminophen, torsemide, albuterol sulfate [ProAir HFA], mycophenolate mofetil, Multivitamin, tacrolimus, Vitamin E, allopurinol, ferrous sulfate.\\\midrule
Triage & 
\begin{tabular}{|l|l|l|l|l|l|l|l|p{1.6in}|}
\centering
temperature &heartrate &resprate &o2sat &sbp &dbp &pain &acuity &chiefcomplaint \\\midrule
98.1 &72.0 &16.0 &0.0 &95.0 &46.0 &8 &2.0 &Abnormal labs, Weakness, Diarrhea \\
\end{tabular}\\
\multicolumn{2}{c}{\cellcolor[RGB]{200,200,200} \textit{Radiologist}} \\
Findings & A new central venous catheter terminates in the left brachiocephalic vein. There is no pneumothorax. Otherwise, there has been no significant short-term change.\\\midrule
Impression & Status post placement of new left internal jugular central venous catheter; no pneumothorax identified.\\
\multicolumn{2}{c}{\cellcolor[RGB]{200,200,200} \textit{Images + effective sources ($h=0$)} ($\textrm{GREEN}=0.143$)} \\
Findings & There is interval placement of a left internal jugular central venous catheter with tip terminating in the lower SVC. Lung volumes are low. This accentuates the size of the cardiac silhouette which appears mildly enlarged. Mediastinal and hilar contours are unchanged. There is crowding of the bronchovascular structures without overt pulmonary edema. Streaky opacities are noted in the lung bases, likely reflective of atelectasis. No large pleural effusion or pneumothorax is demonstrated. Mild degenerative changes are noted in the thoracic spine.\\\midrule
Impression & Interval placement of a left internal jugular central venous catheter with tip in the lower SVC. Low lung volumes with streaky bibasilar opacities, likely atelectasis.\\
\multicolumn{2}{c}{\cellcolor[RGB]{200,200,200} \textit{Images} ($\textrm{GREEN}=0.25$)} \\
Findings & A PICC line terminates in the mid-to-lower SVC. The cardiomediastinal and hilar contours are within normal limits. The lung fields are clear. There is no pneumothorax, fracture or dislocation. Limited assessment of the abdomen is unremarkable.\\\midrule
Impression & Left PICC terminates in the mid-to-lower SVC.\\
\bottomrule
\end{tabular}
\end{table*}

\subsubsection{False Positive: Example 2} \label{sec:fp_eg2}

Table \ref{tab:54082940} presents a case where the model using auxiliary patient data generated false positive predictions, identifying mild pulmonary vascular congestion and a mildly enlarged cardiac silhouette (cardiomegaly). In contrast, the model without auxiliary patient data did not produce these errors. Shortness of breath, noted in the indication section and the chief complaint from the triage data, is a common symptom of both mild pulmonary vascular congestion and cardiomegaly. The elevated respiratory rate recorded in the triage data, consistent with tachypnoea, may suggest mild pulmonary vascular congestion, while the elevated systolic blood pressure reflects isolated systolic hypertension, a known risk factor for cardiomegaly. Furosemide, included in the reconciled medicines, can help manage mild pulmonary vascular congestion associated with fluid overload and conditions like cardiomegaly. Lisinopril and diltiazem primarily treat hypertension, which is a risk factor for cardiomegaly. This example indicates that there was evidence in the auxiliary patient data that could have led the model to multiple false positive predictions. For this example, the model lacked the ability to correctly balance auxiliary patient data evidence with imaging evidence.

\begin{table*}[t]
\centering
\caption{\label{tab:54082940} False positive example for study \texttt{54082940}. This example demonstrates how the model failed to balance auxiliary patient data evidence with imaging evidence. Only the patient data that \textit{Images + effective sources (h=0)} utilises is shown.}
\small
\setlength{\tabcolsep}{5pt} % Reduce column spacing
\begin{tabular}{p{0.5in}p{5.6in}}
\toprule
\multicolumn{2}{c}{\cellcolor[RGB]{200,200,200} \textit{Patient data}} \\
Image & \includegraphics[scale=0.4, valign=c]{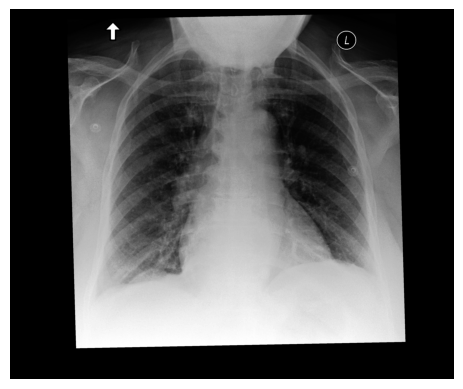} \includegraphics[scale=0.4, valign=c]{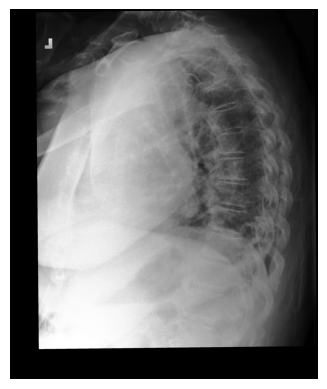} \\\midrule
Indication & Shortness of breath and wheezing, previously diagnosed with pneumonia or infectious process. \\\midrule
Reconciled medicines; name & prednisolone acetate, albuterol sulfate [ProAir HFA], gabapentin, Humulin 70/30, cholecalciferol (vitamin D3), sennosides [senna], furosemide, Trusopt, lisinopril, AERO CHAMBER, levobunolol, insulin aspart, insulin aspart [Novolog], fluticasone-salmeterol [Advair Diskus], latanoprost, dorzolamide [Trusopt], aspirin [Enteric Coated Aspirin], diltiazem HCl [DILT-XR], blood sugar diagnostic [FreeStyle Lite Strips], magnesium hydroxide [Milk of Magnesia], Acetaminophen, lidocaine, docusate sodium, brimonidine, calcium carbonate, nebulizer and compressor, olanzapine [Zyprexa].\\\midrule
Triage & 
\begin{tabular}{|l|l|l|l|l|l|l|l|l|}
\centering
temperature &heartrate &resprate &o2sat &sbp &dbp &pain &acuity &chiefcomplaint \\\midrule
98.0 &81.0 &24.0 &100.0 &151.0 &66.0 &0 &2.0 &SHORTNESS OF BREATH \\
\end{tabular}\\
\multicolumn{2}{c}{\cellcolor[RGB]{200,200,200} \textit{Radiologist}} \\
Findings & There is no evidence of focal consolidation. There is left lower lobe atelectasis. There is no pleural effusion or pneumothorax. The cardiac and mediastinal contours are normal.\\\midrule
Impression & No acute cardiopulmonary process.\\
\multicolumn{2}{c}{\cellcolor[RGB]{200,200,200} \textit{Images + effective sources ($h=0$)} ($\textrm{GREEN}=0.429$)} \\
Findings & There is mild pulmonary vascular congestion. No definite focal consolidation is seen. No pleural effusion or pneumothorax is seen. Cardiac silhouette is mildly enlarged. The cardiac and mediastinal silhouettes are grossly stable with the cardiac silhouette possibly slightly enlarged compared to prior.\\\midrule
Impression & Mild pulmonary vascular congestion. Cardiomegaly.\\
\multicolumn{2}{c}{\cellcolor[RGB]{200,200,200} \textit{Images} ($\textrm{GREEN}=0.8$)} \\
Findings & There is no confluent consolidation. No pulmonary edema or pleural effusions are identified. Cardiomediastinal and hilar contours are within normal limits. No pneumothorax is evident.\\\midrule
Impression & No acute cardiopulmonary process.\\
\bottomrule
\end{tabular}
\end{table*}

\subsubsection{True Negative: Example 1} \label{sec:tn_eg1}

Table \ref{tab:52428322} shows an exam where the model using auxiliary patient data generated a report with true negatives, despite confounding evidence from the auxiliary patient data. The renal failure and upper quadrant pain mentioned in the history section could suggest a condition related to fluid overload, such as pleural effusion. Furosemide and metolazone mentioned in the reconciled medicines are commonly used for fluid management and treating pulmonary edema. Lisinopril and amlodipine, primarily used for cardiovascular conditions such as hypertension, can lead to secondary effects like pulmonary congestion or cardiomegaly, which may be detected radiologically. Despite these confounding factors, the model effectively prioritised the imaging evidence, avoiding false positive predictions. This demonstrates that the model possesses the ability to balance auxiliary patient data evidence with imaging evidence.

\begin{table*}[t]
\centering
\caption{\label{tab:52428322} True negative example for study \texttt{52428322}. This demonstrates how the model can avoid false positives despite confounding evidence from the auxiliary patient data. Only the patient data that \textit{Images + effective sources (h=0)} utilises is shown.}
\small
\setlength{\tabcolsep}{5pt} % Reduce column spacing
\begin{tabular}{p{0.5in}p{5.6in}}
\toprule
\multicolumn{2}{c}{\cellcolor[RGB]{200,200,200} \textit{Patient data}} \\
Image & \includegraphics[scale=0.4, valign=c]{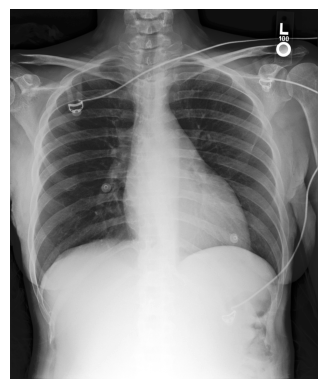} \\\midrule
History & \_\_\_-year-old female with renal failure and right upper quadrant pain. Hypotension. \\\midrule
Reconciled medicines; name & aspirin, FreeStyle Lite Strips, metolazone, furosemide, omeprazole, oxycodone-acetaminophen [Endocet], prochlorperazine maleate, calcitriol, fentanyl, insulin glargine [Lantus], sevelamer HCl [Renagel], clonidine [Catapres-TTS-1], Humulin R, Colace, insulin lispro [Humalog], potassium chloride [Klor-Con M20], FreeStyle Lite Meter, ferrous sulfate, lisinopril, BD Insulin Syringe Ultra-Fine, Glucose Meter, Disp \& Strips, Lipitor, amlodipine. \\\midrule
Triage & 
\begin{tabular}{|l|l|l|l|l|l|l|l|l|}
\centering
temperature &heartrate &resprate &o2sat &sbp &dbp &pain &acuity &chiefcomplaint \\\midrule
0.0 &0.0 &0.0 &0.0 &0.0 &0.0 &None &1.0 &GI BLEED \\
\end{tabular}\\
\multicolumn{2}{c}{\cellcolor[RGB]{200,200,200} \textit{Radiologist}} \\
Findings & Single portable view of the chest is compared to previous exam from \_\_\_. The lungs are clear. Cardiomediastinal silhouette is normal. Osseous and soft tissue structures are unremarkable. No visualized free intraperitoneal air is seen below the diaphragm.
 \\\midrule
Impression & No acute cardiopulmonary process. \\
\multicolumn{2}{c}{\cellcolor[RGB]{200,200,200} \textit{Images + effective sources ($h=0$)} ($\textrm{GREEN}=1.0$)} \\
Findings & PA and lateral views of the chest were provided. The lungs are clear bilaterally without focal consolidation, effusion, or pneumothorax. The heart and mediastinal contours are normal. The imaged osseous structures are intact. There is no free air below the right hemidiaphragm. \\\midrule
Impression & No acute findings in the chest. \\
\multicolumn{2}{c}{\cellcolor[RGB]{200,200,200} \textit{Images} ($\textrm{GREEN}=0.5$)} \\
Findings & Single portable frontal chest radiograph demonstrates unremarkable cardiomediastinal and hilar contours. Lungs are clear. No pleural effusion or pneumothorax evident. \\\midrule
Impression & No acute intrathoracic process.\\
\bottomrule
\end{tabular}
\end{table*}

\subsubsection{True Negative: Example 2} \label{sec:tn_eg2}

Table \ref{tab:52169517} is another exam where the model using auxiliary patient data generated a report with true negatives, despite confounding evidence from the auxiliary patient data. The model utilising auxiliary patient data accurately identified a dual-lead pacemaker without introducing any false positive findings, despite the presence of confounding evidence from the auxiliary patient data. The indication section requests evaluation for fluid overload or pneumonia and notes chest pain, which could lead to false positives such as pulmonary edema, pneumonia, pleural effusion, or cardiomegaly. The reconciled medicines, including furosemide and nitroglycerin, suggest the management of conditions such as pulmonary edema or heart failure, which could be associated with pleural effusion or cardiomegaly. Despite these confounding factors, the model effectively prioritised the evidence from the image, avoiding false positive predictions. 

\begin{table*}[t]
\centering
\caption{\label{tab:52169517} True negative example for study \texttt{52169517}. This demonstrates how the model can avoid false positives despite confounding evidence from the auxiliary patient data. Only the patient data that \textit{Images + effective sources (h=0)} utilises is shown.}
\small
\setlength{\tabcolsep}{5pt} % Reduce column spacing
\begin{tabular}{p{0.5in}p{5.6in}}
\toprule
\multicolumn{2}{c}{\cellcolor[RGB]{200,200,200} \textit{Patient data}} \\
Image & \includegraphics[scale=0.4, valign=c]{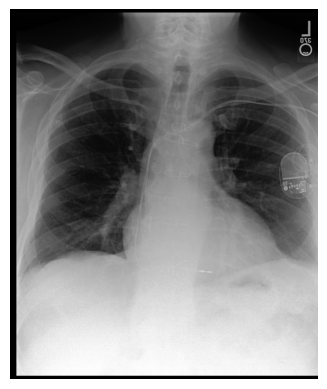}\includegraphics[scale=0.4, valign=c]{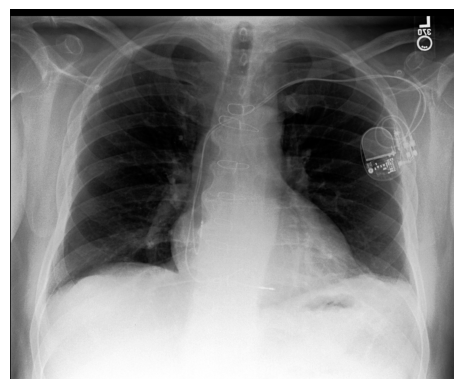}\includegraphics[scale=0.4, valign=c]{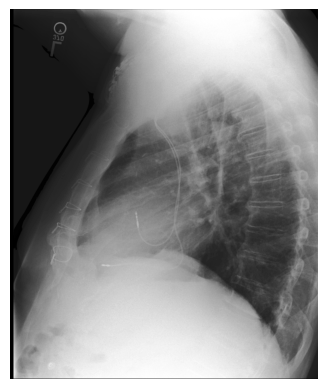} \\\midrule
Indication & \_\_\_-year-old woman with chest pain. Evaluate for fluid overload or pneumonia. \\\midrule
Reconciled medications; name & Humalog, atorvastatin, aspirin, gabapentin, nitroglycerin, methylprednisolone, valsartan [Diovan], insulin glargine [Lantus], One Touch Ultra Test, metoprolol tartrate, isosorbide mononitrate, cephalexin, colchicine [Colcrys], furosemide \\\midrule
Triage & 
\begin{tabular}{|l|l|l|l|l|l|l|l|l|}
\centering
temperature &heartrate &resprate &o2sat &sbp &dbp &pain &acuity &chiefcomplaint \\\midrule
0.0 &0.0 &0.0 &0.0 &0.0 &0.0 &None &2.0 &CHEST PAIN \\
\end{tabular}\\
\multicolumn{2}{c}{\cellcolor[RGB]{200,200,200} \textit{Radiologist}} \\
Findings & The lungs are clear. The hilar and cardiomediastinal contours are normal. There is no pneumothorax or pleural effusion. Pulmonary vascularity is normal. A dual-lead pacemaker is present.\\\midrule
Impression & No acute cardiopulmonary process.\\
\multicolumn{2}{c}{\cellcolor[RGB]{200,200,200} \textit{Images + effective sources ($h=0$)} ($\textrm{GREEN}=0.833$)} \\
Findings & The lungs are clear without focal consolidation. No pleural effusion or pneumothorax is seen. The cardiac and mediastinal silhouettes are unremarkable. Dual lead left-sided pacemaker is stable in position. Patient is status post median sternotomy and CABG.\\\midrule
Impression & No acute cardiopulmonary process.\\
\multicolumn{2}{c}{\cellcolor[RGB]{200,200,200} \textit{Images} ($\textrm{GREEN}=0.25$)} \\
Findings & PA and lateral views of the chest provided demonstrate a left chest wall pacer device with leads extending to the region of the right atrium and right ventricle. Midline sternotomy wires and mediastinal clips are again noted. Mild left basal atelectasis is noted. No convincing signs of pneumonia, effusion, or pneumothorax. The cardiomediastinal silhouette is stable. Bony structures are intact. No free air below the right hemidiaphragm.\\\midrule
Impression & No acute findings in the chest.\\
\bottomrule
\end{tabular}
\end{table*}

\subsubsection{False Negative: Example 1} \label{sec:fn_eg1}

Table \ref{tab:55715754} is an example where the model failed to leverage auxiliary patient data to detect trace bilateral pleural effusions and the increased opacity in the right mid-to-lower lung (concerning for pneumonia). The history section notes dyspnea and hypoxia, which are symptoms associated with pleural effusion and pneumonia, among other conditions. It also requests to assess for fluid overload or pneumonia, both of which should prompt the model to assess for pleural effusion and opacities. The significantly reduced oxygen saturation recorded in the triage data indicates severe hypoxia (also noted in the history section), which can be caused by pleural effusion or pneumonia. Despite strong evidence from the auxiliary patient data to support pleural effusion and the opacity, the model failed to combine this with the imaging evidence to make the correct predictions.

\begin{table*}[t]
\centering
\caption{\label{tab:55715754} False negative example for study \texttt{55715754}. The model failed to identify the pleural effusions despite evidence from the auxiliary patient data. Only the patient data that \textit{Images + effective sources (h=0)} utilises is shown.}
\small
\setlength{\tabcolsep}{5pt} % Reduce column spacing
\begin{tabular}{p{0.5in}p{5.6in}}
\toprule
\multicolumn{2}{c}{\cellcolor[RGB]{200,200,200} \textit{Patient data}} \\
Image & \includegraphics[scale=0.4, valign=c]{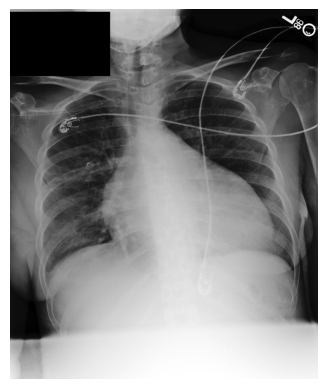} \\\midrule
History & Dyspnea and hypoxia, assess for fluid overload or pneumonia.\\\midrule
Triage & 
\begin{tabular}{|l|l|l|l|l|l|l|l|l|}
\centering
temperature &heartrate &resprate &o2sat &sbp &dbp &pain &acuity &chiefcomplaint \\\midrule
96.4 &83.0 &20.0 &76.0 &145.0 &70.0 &10 &1.0 &SORE THROAT \\
\end{tabular}\\
\multicolumn{2}{c}{\cellcolor[RGB]{200,200,200} \textit{Radiologist}} \\
Findings &Semi-upright portable AP view of the chest provided. The heart is massively enlarged. There are trace pleural effusions. Increased opacity in the right mid-to-lower lung is concerning for pneumonia. The left lung appears essentially clear. No pneumothorax. The mediastinal contour appears normal. Bony structures are intact.\\\midrule
Impression & Massive cardiomegaly with trace bilateral pleural effusions. Opacity within the right mid-to-lower lung is concerning for pneumonia.\\
\multicolumn{2}{c}{\cellcolor[RGB]{200,200,200} \textit{Images + effective sources ($h=0$)} ($\textrm{GREEN}=0.2$)} \\
Findings &Single portable radiograph of the chest demonstrates moderate enlargement of the cardiac silhouette, not significantly changed compared to the prior examination. There is mild pulmonary vascular congestion. No focal consolidation, pleural effusion or pneumothorax is seen. The visualized upper abdomen is unremarkable.\\\midrule
Impression & Persistent enlargement of the cardiac silhouette, not significantly changed compared to \_\_\_. Unchanged mild pulmonary vascular congestion and stable enlargement of the cardiac silhouette.\\
\multicolumn{2}{c}{\cellcolor[RGB]{200,200,200} \textit{Images} ($\textrm{GREEN}=0.333$)} \\
Findings & There is moderate enlargement of the cardiac silhouette. The aorta is unfolded. Mediastinal and hilar contours are otherwise unremarkable. Pulmonary vasculature is not engorged. Hazy opacity in the right lung is compatible with pneumonia. Right midlung linear opacity may be due to atelectasis. No pleural effusion or pneumothorax is identified. No acute osseous abnormalities seen.\\\midrule
Impression & 1. Moderate enlargement of the cardiac silhouette, compatible with pneumonia. 2. Moderate enlargement of the cardiac silhouette. 3. Right lung base opacity, likely scarring. No definite evidence of pneumonia.\\
\bottomrule
\end{tabular}
\end{table*}

\subsubsection{False Negative: Example 2} \label{sec:fn_eg2}

Table \ref{tab:53964812} highlights a false negative example for the model leveraging auxiliary patient data, where it failed to identify the right lower lobe opacity concerning for pneumonia. Despite omitting this finding, the model received strong evidence supporting its presence. Specifically, the indication section notes a right lower lobe infiltrate, directly pointing to an opacity, alongside dyspnoea, a non-specific symptom of pneumonia. Additionally, the chief complaint explicitly lists pneumonia, another strong indicator. The triage data, including the normal temperature and heart rate, might have influenced the model’s decision by suggesting a lack of systemic immune response, which could reduce the likelihood of pneumonia. However, the reconciled medicines, including antibiotics like erythromycin and tobramycin-dexamethasone, support the possibility of an active infection. Despite the alignment between the auxiliary patient data and the suspected pneumonia, the model failed to integrate this evidence with the imaging evidence to make a correct prediction. This underscores the need for further model development to better synthesise the evidence from auxiliary patient data with imaging evidence.

\begin{table*}[t]
\centering
\caption{\label{tab:53964812} False negative example for study \texttt{53964812}. Despite strong evidence from the auxiliary patient data supporting pleural effusion, the model failed to detect it. Only the patient data that \textit{Images + effective sources (h=0)} utilises is shown.}
\small
\setlength{\tabcolsep}{5pt} % Reduce column spacing
\begin{tabular}{p{0.5in}p{5.6in}}
\toprule
\multicolumn{2}{c}{\cellcolor[RGB]{200,200,200} \textit{Patient data}} \\
Image & \includegraphics[scale=0.4, valign=c]{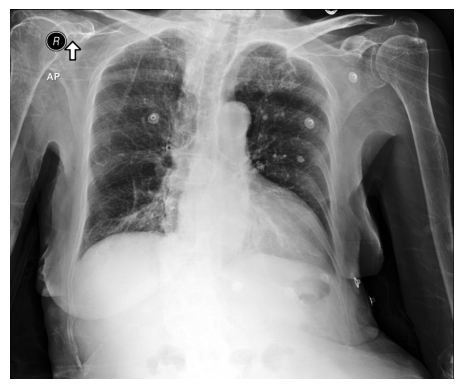}\includegraphics[scale=0.4, valign=c]{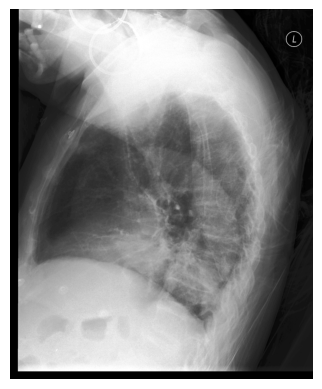} \\\midrule
History & Asthma, coronary disease, myocardial infarction, diabetes, presenting with dyspnea and right lower lobe infiltrate.\\\midrule
Reconciled medicines; name & metformin, acetaminophen, erythromycin, fluticasone-salmeterol [Advair Diskus], Boost Diabetic, bupropion HCl, diltiazem HCl, albuterol sulfate, losartan [Cozaar], lorazepam, sennosides [senna], gabapentin, aspirin, tiotropium bromide [Spiriva with HandiHaler], tobramycin-dexamethasone [TobraDex], atorvastatin, tramadol, nut.tx.gluc.intol,lac-free,soy [Glucerna], clopidogrel, ReliOn Ultra Thin Plus Lancets, melatonin, lidocaine, calcium carbonate-vitamin D3 [Calcium 600 + D(3)], furosemide, compression socks, medium, sertraline, L.acidoph \& sali-B.bif-S.therm [Acidophilus], travoprost [Travatan Z], montelukast [Singulair], Multivitamin with Iron-Mineral, fluticasone.\\\midrule
Triage & 
\begin{tabular}{|l|l|l|l|l|l|l|l|l|}
\centering
temperature &heartrate &resprate &o2sat &sbp &dbp &pain &acuity &chiefcomplaint \\\midrule
98.2 &83.0 &18.0 &98.0 &162.0 &100.0 &0 &2.0 &PNEUMONIA \\
\end{tabular}\\
\multicolumn{2}{c}{\cellcolor[RGB]{200,200,200} \textit{Radiologist}} \\
Findings &T0he cardiac, mediastinal and hilar contours appear stable. There is no pleural effusion or pneumothorax. Since the very recent prior studies, there is a substantial new opacity in the right lower lobe concerning for pneumonia. The bones appear demineralized. There is mild-to-moderate rightward convex curvature again centered along the lower thoracic spine with incompletely characterized lumbar compression deformities. Moderate degenerative changes are again noted along lower thoracic levels.\\\midrule
Impression & Findings consistent with pneumonia in the right lower lobe. Depending on clinical circumstances, the possibility of aspiration could also be considered.\\
\multicolumn{2}{c}{\cellcolor[RGB]{200,200,200} \textit{Images + effective sources ($h=0$)} ($\textrm{GREEN}=0.0$)} \\
Findings & Frontal and lateral views of the chest. Right apical scarring is again seen. The lungs are otherwise clear without consolidation or effusion. Mild cardiomegaly is again noted. Slightly tortuous descending thoracic aorta is similar to prior. No acute osseous abnormality is identified.\\\midrule
Impression& No acute cardiopulmonary process.\\
\multicolumn{2}{c}{\cellcolor[RGB]{200,200,200} \textit{Images} ($\textrm{GREEN}=0.333$)} \\
Findings & There is bibasilar atelectasis without definite focal consolidation. No pleural effusion or pneumothorax is seen. The cardiac and mediastinal silhouettes are stable. Mild loss of height anteriorly of a lower thoracic vertebral body is unchanged. Evidence of DISH is seen along the spine.\\\midrule
Impression & No acute cardiopulmonary process. No significant interval change.\\
\bottomrule
\end{tabular}
\end{table*}

% \section{Example Appendix}
% \label{sec:appendix}

% This is an appendix.

\end{document}